\documentclass[11pt]{article}
\usepackage[utf8]{inputenc}
\usepackage{amsfonts}

\usepackage[final]{acl}
\usepackage{times}
\usepackage{pgfplots}
\usepackage{tikz}
\usepackage{amsmath, amssymb}
\usepackage{latexsym}

\usepackage{algorithmicx}
\usepackage{algorithm}

\usepackage{xcolor}
\usepackage{tcolorbox}
\tcbuselibrary{skins, breakable}
\usepackage{listings}
\usepackage{lipsum}
\usepackage{multicol}
\usepackage{multirow}
\usepackage{booktabs} % \toprule, \midrule, \bottomrule
\usepackage{arydshln} % \hdashline
\usepackage{subcaption} % subfigure environment
\usepackage{float} % [H] specifier

\usepackage[noend]{algpseudocode} % [noend] 保证没有 "end if", "end for" 等

\usepackage[T1]{fontenc}
\usepackage[utf8]{inputenc}

\usepackage{microtype}
\usepackage{inconsolata}

\usepackage{graphicx}

\definecolor{darkblue}{HTML}{000099}
\definecolor{darkpurple}{HTML}{4B0082}

\hypersetup{
    colorlinks=true,       % 开启彩色链接（这一步会自动去掉默认的丑陋红框/绿框）
    linkcolor=darkblue,    % 文档内部链接（如 Figure 1, Table 2, Section 3 的跳转）
    urlcolor=darkblue,     % 外部 URL 链接（如 Github 网址）
    citecolor=darkblue,    % 参考文献引用颜色（如 [1], Zhao et al.）
    anchorcolor=darkblue,  % 锚点颜色
    pdfborderstyle={/S/U/W 0} % 加上这句作为双重保险，彻底消除任何边框或下划线
}
\makeatletter
\def\blfootnote{\gdef\@thefnmark{}\@footnotetext}
\makeatother

\title{STEM: Structure-Tracing Evidence Mining for Knowledge Graphs-Driven Retrieval-Augmented Generation}

\author{
    \textbf{Peng Yu}, \textbf{En Xu}, \textbf{Bin Chen}\thanks{Corresponding author.}, \textbf{Haibiao Chen}, \textbf{Yinfei Xu} \\
    AI Product Center, Kingsoft Corporation, Beijing, China \\
    \texttt{\textcolor{darkpurple}{\{yupeng5,xuen,chenbin,chenhaibiao,xuyinfei\}@kingsoft.com}} \\
}

\begin{document}
\maketitle
\blfootnote{Accepted to the Main Conference of ACL 2026 (Oral Presentation).}
\begin{abstract}
Knowledge Graph-based Question Answering (KGQA) plays a pivotal role in complex reasoning tasks but remains constrained by two persistent challenges: the structural heterogeneity of Knowledge Graphs (KGs) often leads to semantic mismatch during retrieval, while existing reasoning path retrieval methods lack a global structural perspective. To address these issues, we propose Structure-Tracing Evidence Mining (STEM), a novel framework that reframes multi-hop reasoning as a schema-guided graph search task. First, we design a Semantic-to-Structural Projection pipeline that leverages KG structural priors to decompose queries into atomic relational assertions and construct an adaptive query schema graph. Subsequently, we execute globally-aware node anchoring and subgraph retrieval to obtain the final evidence reasoning graph from KG. To more effectively integrate global structural information during the graph construction process, we design a Triple-Dependent GNN (Triple-GNN) to generate a Global Guidance Subgraph (Guidance Graph) that guides the construction. STEM significantly improves both the accuracy and evidence completeness of multi-hop reasoning graph retrieval, and achieves State-of-the-Art performance on multiple multi-hop benchmarks. Our source code is available at \url{https://github.com/PennyYu123/STEM_RAG}.
\end{abstract}

\section{Introduction}
\label{sec:intro}
The research and development of large language models (LLMs) have spanned several years \cite{openai2023gpt4,anthropic2024claude,touvron2023llama,chowdhery2023palm,jiang2023mistral,bai2023qwen}, however, LLMs suffer from the issue of hallucination, often leading to inaccuracies in responses to fact-based questions. To address this, Retrieval-Augmented Generation (RAG) was introduced as a promising paradigm to ground model outputs in verifiable external knowledge bases\cite{lewis2020retrieval,trivedi2022interleaving,guu2020realm,borgeaud2022improving}. By leveraging pre-existing knowledge bases, RAG enables LLMs to reference relevant contextual information when generating answers, thereby improving the accuracy and quality of responses. In recent years, knowledge graph-based question answering systems for LLMs have gained significant research attention \cite{yasunaga2021qa,zhang2022greaselm,he2024gretriever,edge2024graphrag}. These systems utilize structured knowledge graphs to organize relationships among various real-world entities, providing the LLM with a hierarchical and logically clear knowledge network, thereby enabling more precise guidance for generating accurate answers. 

\begin{figure}[t]
    \centering
    \includegraphics[width=\columnwidth, trim=2.5pt 2.5pt 2.5pt 2.5pt, clip]{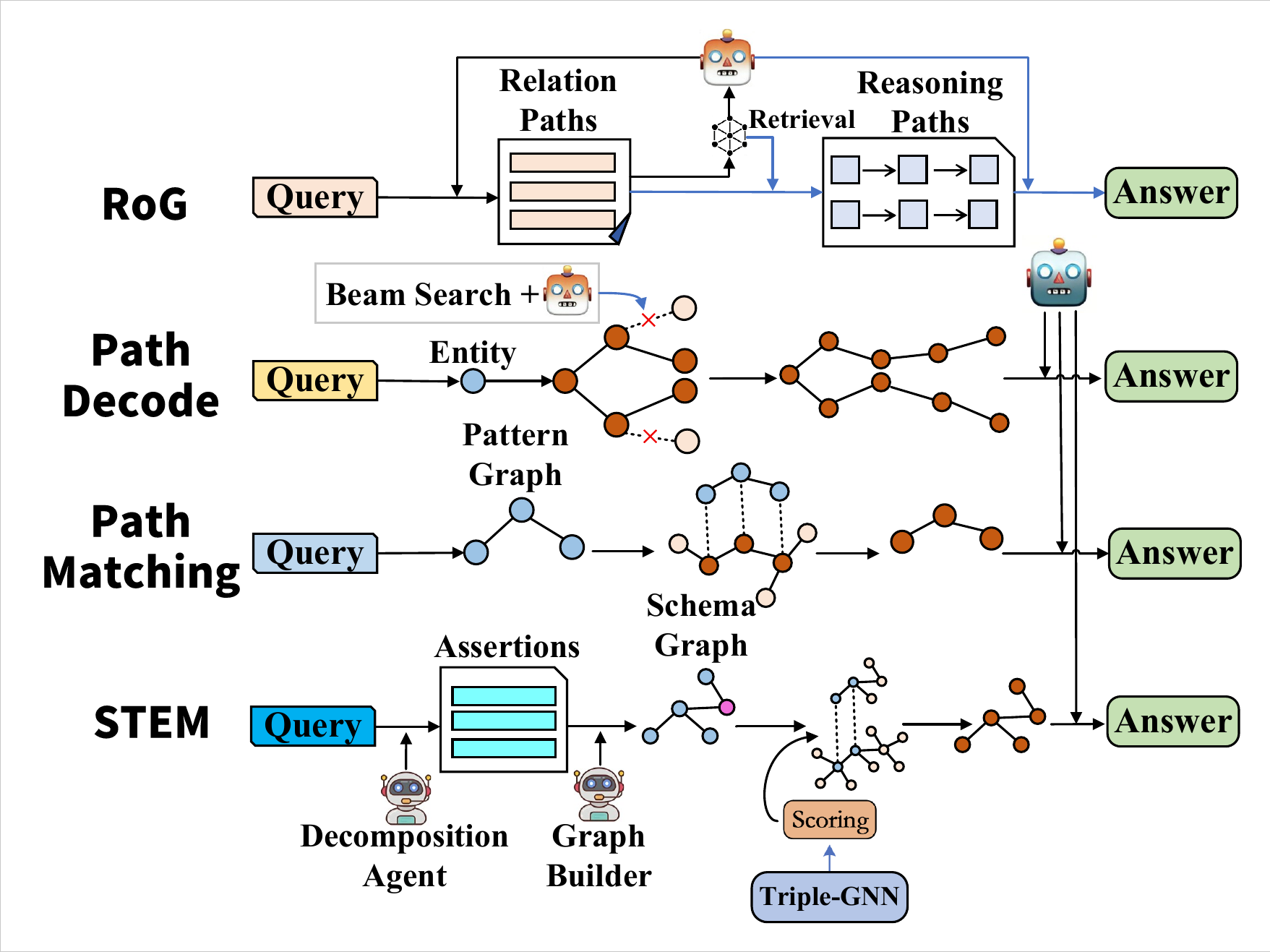}
    \caption{Different KG Retrieval Reasoning Frameworks.}
    \label{fig:overview}
\end{figure}

Knowledge graph-based multi-hop reasoning for question answering has also garnered extensive research attention \cite{yasunaga2021qa,luo2024reasoning,yu2023decaf,sui-etal-2025-fidelis,liu2025pathmind,cai2025simgrag}. Existing approaches for KG-based reasoning generally fall into three main paradigms: The first involves generating reasoning plans based on the query to retrieve evidence chains for answer generation (RoG) \cite{luo2024reasoning}. The second employs step-wise path exploration, such as beam search, utilizing intelligent modules like LLMs to iteratively determine the optimal path (Path Decode) \cite{sui-etal-2025-fidelis}. The third focuses on structural matching, which involves constructing an aligned schema graph to guide step-by-step searches within the KG to build a minimal distance subgraph (Path Matching) \cite{cai2025simgrag}. The retrieved subgraph is subsequently verbalized and fed into an LLM for final answer generation. An overview of the design of the above method is detailed in Figure \ref{fig:overview}.

Despite the extensive research in multi-hop reasoning over Knowledge Graphs (KGs), existing methods still face significant impediments that limit their efficacy and robustness. We identify three primary challenges as follows:

\textbf{First, the semantic-structural gap between LLM-generated plans and KG schemas hinders accurate retrieval.} KGs are characterized by inherent organizational complexity, containing millions of entities and diverse relation types. Current approaches typically rely on LLMs to decompose questions or generate reasoning plans in natural language. However, due to the LLM's lack of prior knowledge regarding the specific KG schema, these generated plans often suffer from \textbf{schema hallucination}—predicting relations that are semantically plausible but topologically non-existent in the target KG. For instance, consider the query ``Which airport to fly into rome?'' While the ground-truth path in the KG follows an hierarchical structure like
``Rome $\xrightarrow{\text{location.location.nearby\_airports}}$ Ciampino–G. B. Pastine International Airport'', but a schema-agnostic planner is likely to generate a forward-looking formulation such as ``[ENT1] $\xrightarrow{\text{fly into}}$ Rome''. This creates a fundamental topological mismatch: the semantic implication of ``fly into'' often fails to align with the knowledge schema, resulting in retrieval failures.  

\textbf{Second, prevailing path-search paradigms suffer from lack of global view and evidence fragmentation.} Most existing methods employ stepwise search strategies, which rely on local semantic similarity or LLM-based decision-making to select the next hop. While some incorporate look-ahead scoring to anticipate future steps, they fundamentally lack a global structural blueprint. This leads to three critical issues: (1) Path Deviation: Without global guidance, the search process is highly sensitive to local spurious semantic correlations. (2)Hub Node Problem: The sheer volume of neighboring relations creates an overwhelming candidate space, which significantly increases the error rate of selection. This necessitates a schema graph for effective search space pruning. (3) Information Fragmentation: Since complex questions often require a subgraph structure rather than isolated reasoning paths, path-based methods frequently retrieve fragmented evidence, consequently the question cannot be adequately answered.

\textbf{Finally, the reliance on interactive reasoning incurs prohibitive computational costs.} Many state-of-the-art methods adopt an ``interleaved'' approach, where the LLM is invoked at every step of the reasoning path to judge validity or refine the search direction, these approaches create a significant bottleneck in inference latency and resource consumption. This bottleneck is particularly exacerbated in multi-answer scenarios, where retrieving multiple answers necessitates traversing parallel reasoning paths, thereby leading to substantially higher LLM overhead. 

To address these challenges, we propose Structure-Tracing Evidence Mining (STEM), a novel architecture that reframes multi-hop KG reasoning from sequential path finding to holistic schema-guided subgraph matching. STEM distinguishes itself through two key innovations: First, we bridge the schema gap via building a Semantic-to-Structural Projection pipeline. Different from existing works attempting to convert queries into logic forms (e.g., SPARQL) for KG retrieval \cite{sun2020sparqa,lan2020query,ye2022rng}, we design two specialized modules: the Schema-Grounded Decomposition Agent (SGDA) and the Symbol-Aligned Graph Builder (SAGB) to ensure the retrieval plan aligns with the KG's inherent topology. The SGDA first decomposes the complex query into a sequence of atomic relational assertions, stripping away linguistic ambiguity and align the semantic narratives with the logic of the KG. Subsequently, the SAGB grounds each assertion into a concrete triple structure, assembling them into a coherent schema graph. 

Second, we implement Structure-Tracing Subgraph Retrieval to retrieve related evidence. STEM employs a Triple-Dependent GNN (Triple-GNN) to produce a Global Guidance Subgraph (Guidance Graph). During the traversal phase, the retrieval is guided not only by local transition probabilities (semantic distance) but also by the global structural scores according to Guidance Graph. This mechanism ensures that search step is globally weighted, effectively correcting potential deviations and ensuring the retrieval of a complete, logically connected evidence subgraph. Based on the comprehensive introduction, the main contributions of this paper are summarized as follows:
\begin{itemize}
    \item We propose a novel Semantic-to-Structural Projection pipeline to bridge the inherent gap between natural language queries and KG schemas. By employing a Schema-Grounded Decomposition Agent and a Symbol-Aligned Graph Builder, we construct a precise schema graph that serves as a topological blueprint for subsequent retrieval.
    \item We introduce Structure-Tracing Evidence Mining, a holistic subgraph matching paradigm for KGQA. By leveraging a Triple-Dependent GNN to construct a Global Guidance Subgraph, our method incorporates global structural priors into the search process, effectively ensuring both the accuracy and completeness of the retrieved evidence.
    \item Our proposed method achieves State-of-the-Art performance on complex multi-hop reasoning benchmarks including WebQSP and CWQ, demonstrating the effectiveness of our proposed method.
\end{itemize}

\section{Preliminaries}
\label{subsec:preliminaries}
Typically, the structure of a knowledge graph can be defined as a tuple: \( \mathcal{G} = (\mathcal{V}, \mathcal{E}, \mathcal{T}, \mathcal{N}, \mathcal{R}) \), where \( \mathcal{V} \) represents the set of nodes, \( \mathcal{E} \) denotes the set of edges, \( \mathcal{T} = \{ (h, r, t) \mid h, t \in \mathcal{V}, r \in \mathcal{E} \} \) represents the set of triplets. \( \mathcal{N} \) corresponds to the textual descriptions of each entity node in \( \mathcal{V} \), i.e., for each node \( v_i \in \mathcal{V} \), there exists \( e_i \in \mathcal{N} \) representing the textual description of \( v_i \). Likewise \( \mathcal{R} \) denotes the set of relations, corresponding to the descriptions of edge in \( \mathcal{E} \), i.e., for each edge \( d_i \in \mathcal{E} \), there exists \( r_i \in \mathcal{R} \) representing the textual description of \( d_i \).

The primary objective of KGQA framework is to identify and extract a relevant subgraph \( \mathcal{G'} \subseteq \mathcal{G} \) from the entire knowledge graph based on a given natural language query $Q$, this retrieved subgraph serves as structured evidence to ground the reasoning process of a Large Language Model. We pre-index all entities mapped to nodes in graph \( \mathcal{G} \) within a high-speed indexing system for the rapid retrieval.

\section{Methodology}

\begin{figure*}[t]
    \centering
    \includegraphics[width=\textwidth, trim=2.5pt 2.5pt 2.5pt 2.5pt, clip]{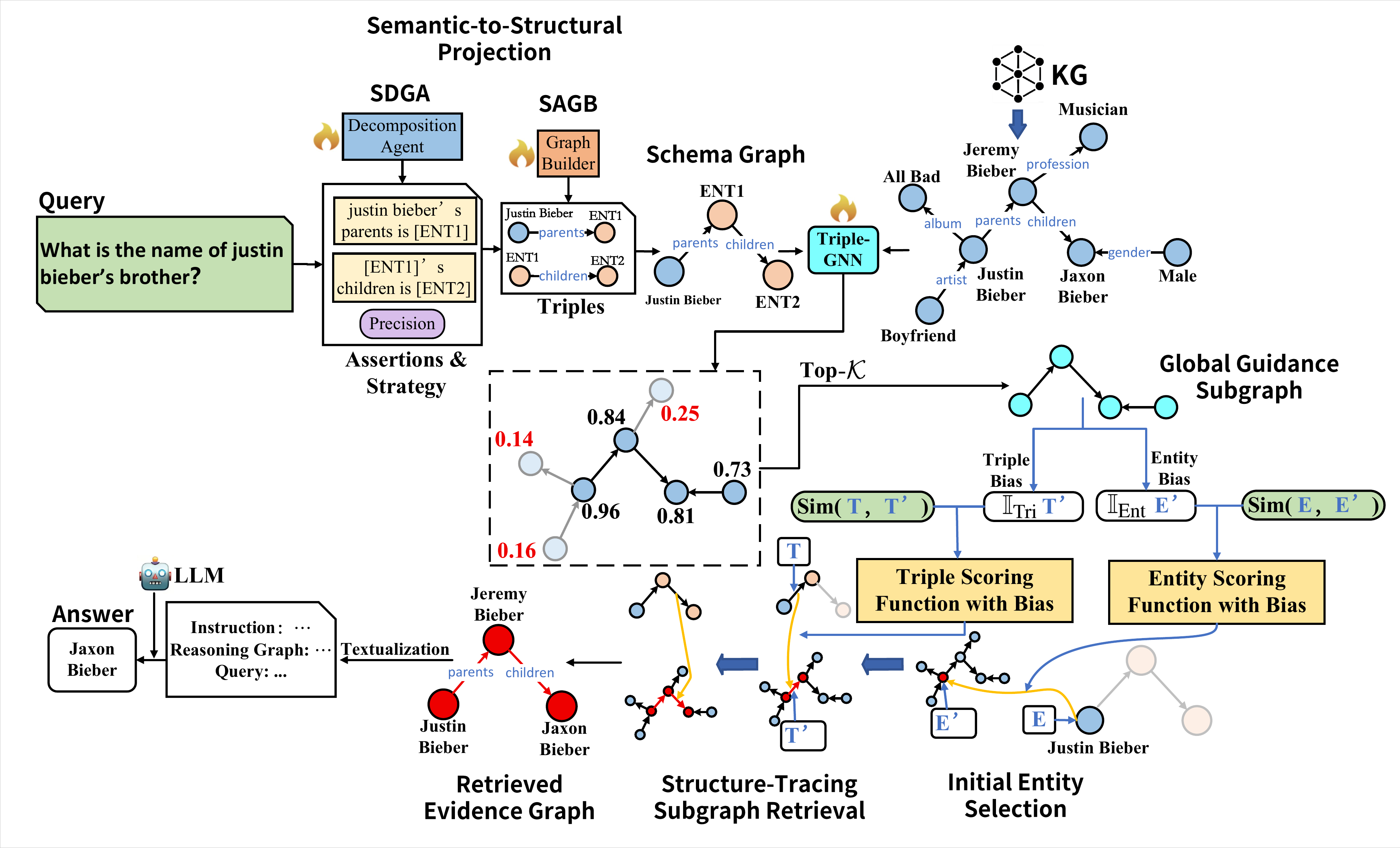}
    \caption{Overview of the STEM Framework.}
    \label{fig:framework}
\end{figure*}

We present our methodology as follows. First, we describe how the SGDA module performs question decomposition. Next, we explain how the SAGB module builds a schema graph. We then introduce the Triple-GNN for Guidance Graph building. Subsequently, we present the complete workflow of Structure-Tracing Subgraph Retrieval. Finally, we introduce the answer generation process. The overall design framework of the approach in this paper is shown in Figure \ref{fig:framework}.

\subsection{Schema-Aligned Question Decomposition}
\label{subsec:SGDA}

The objective of the SGDA is to generate atomic relational assertions from the original query, these assertions will align the semantic narratives with the relations in KG. CoG \cite{zhao-etal-2025-correcting} proposed a ``knowledge reciting'' task, training an LLM with query-path pairs from datasets to equip with prior knowledge of paths, preventing plausible but irretrievable paths in the KG. Inspired by this, and considering the scale and complexity of nodes and paths in KGs, our SGDA module focuses on learning patterns rather than specific knowledge. For instance, given the training example (``What is the San Francisco Giants' mascot?'', ``San Francisco Giants' mascot is [ENT1]''), the SGDA can construct assertions following the same pattern for similar queries, such as ``What is the X's mascot?'', successfully resolving cases associated with this type of problem.

\textbf{Atomic Relational Assertion.} An atomic relational assertion refers to a text describing a single logical relationship—for instance, ``Western Sahara contains Smara''—which can be directly mapped to a single triple (Western Sahara, contains, Smara). Formally, Given the prompt \hyperref[fig:p1]{$ \mathcal{P}_1 $} and the multi-hop reasoning  question $Q$, SGDA decompose the original query into a structured sequence \( \mathcal{S} \), where each element is represented as an assertion $s$, the form is described below: 
\begin{align}
\mathcal{S} = \{ s_1, s_2, , ... , s_N \}
\end{align}

We present the specific content of $\mathcal{P}_1$ and all subsequent prompts in the Appendix \ref{sec:promptlist}. To maintain logical connectivity across multi-hop reasoning steps, the SGDA employs a unified placeholder mechanism for entity linking. Since the answer to a preceding sub-question often serves as the bridging entity for the subsequent one, we standardize these intermediate variables using shared identifiers (e.g., [ENTX]). For instance, given the multi-hop query ``Where is the arena stadium of the team whose mascot is Clutch the Bear?'', the SGDA decomposes it into a coherent sequence of assertions sharing the bridging entity [ENT1]:
\begin{align*}
1. & \text{ENT1}\text{'s mascot is Clutch the Bear} \\
2. & \text{ENT1}\text{'s arena stadium is [ENT2]}
\end{align*}

\textbf{Answer Strategy.} We consider multi-answer scenarios. For instance, the question ``what are the four main languages spoken in Spain'' corresponds to multiple potential answers, retrieving only a single evidence graph will result in incomplete answers. To address this, we categorize questions into two types: \textit{Precision} and \textit{Breadth}. The former corresponds to a definitive answer, while the latter involves multiple valid answers. The distinction between these two types will influence the behavior of subgraph matching, as detailed in subsequent sections. Since a single query may correspond to diverse potential KG logical structures, we employ beam search to enable the SGDA to generate \( \mathcal{B} \) multiple candidate results, finally constructing a list of candidate assertion-strategy pairs $(\mathcal{S}_Q, \sigma)$ corresponding to different planning results, $\sigma$ signifies the retrieval strategy assigned to $Q$,where $\sigma \in \{``Precision", ``Breadth"\}$.

\subsection{Symbol-Aligned Graph Construction}
\label{subsec:SAGB}

While the SGDA successfully decomposes complex queries into atomic relational assertions, the objective of the SAGB is to perform symbolic grounding: mapping these textual assertions into precise structural triples. For instance, consider the assertion: ``Darryl Sutter's hockey position is [ENT1].'', A naive keyword match might fail to identify the correct edge due to lexical divergence (e.g., (``Darryl Sutter'', ``position'', ``[ENT1]'')). However, possessing prior knowledge of the KG's symbolic representation, the SAGB accurately grounds this assertion into the standardized triple: (``Darryl Sutter'', ``ice\_hockey.hockey\_player.hockey\_position'', ``[ENT1]'').

Formally, for a given set of assertions and the prompt \hyperref[fig:p2]{$ \mathcal{P}_2 $} , denoted as \(\mathcal{S} = \{ s_1, s_2, , ... , s_N \}\), SAGB build a corresponding set of structural triples, represented as
\begin{align}
\mathcal{T} = \{ t_1, t_2, , ... , t_N \}
\end{align}

After obtaining the set of triples $\mathcal{T}$, we construct the schema graph $\mathcal{G}_{sch}$ by traversing each member of $\mathcal{T}$. 

To develop these two modules, we propose a specialized data construction and training framework. Furthermore, to enhance the KG knowledge injection and generalization capabilities of the SGDA and SAGB, we introduce Structure-to-Query Reverse Generation method for data augmentation. All details will be described in Appendix \ref{sec:training}. 

In Appendix \ref{sec:projection_pipeline}, we provide concrete running examples to illustrate the end-to-end processing workflow of the Semantic-to-Structural Projection pipeline and the underlying principles of pattern acquisition during training.

\subsection{Global Guidance Subgraph}
\label{subsec:initial_graph_select}

Graph Neural Networks (GNNs) have been widely adopted for reasoning over KGs \cite{mavromatis-karypis-2025-gnn,yasunaga2021qa,liu2025pathmind}. GFM-RAG \cite{luo2025gfmrag_arxiv} typically employs a Query-Dependent GNN, which incorporates query information to compute interaction-aware representations, enabling the ability to capture relevant entity knowledge within the graph structure. Based on this approach, we propose the Triple-Dependent GNN (Triple-GNN), which leverages the explicit structural relationship inherent in triples.

Formally, let \( \mathcal{G} \) denote the query-specific subgraph\footnote{We utilize the subgraph structures extracted by RoG \cite{luo2024reasoning}. In their publicly available dataset, each query corresponds to a specific KG and a list of question entities, all of which are derived from Freebase.} corresponding to $Q$, which serves as the knowledge graph for retrieval. $ \mathcal{N} $ and $ \mathcal{R} $ denote the set of entities and relations within \( \mathcal{G} \), respectively, and  \(\mathcal{N}_Q\) denotes the set of all question entities in $Q$. After obtaining the structural triples \(\mathcal{T}\) and schema graph $\mathcal{G}_\text{sch}$. For each triple \(t \in \mathcal{T}\), we first employ a Pretrained Embedding Model (PEM) to generate its vector representation $\mathbf{E}_{t} \in \mathbb{R}^{d_\text{GNN}}$. A pooling operation is then applied to integrate all $\mathbf{E}_{t_i}$ ($t_i \in$ \( \mathcal{T} \)) into a unified representation $\mathbf{E}_Q \in \mathbb{R}^{d_\text{GNN}}$. In the message passing stage, $\mathbf{E}_Q$ is fed into the $L$-layer Triple-GNN to derive the embedding representation $\mathbf{h}^L_e$ for each entity $e \in \mathcal{N}_Q$. We consolidate all entity representations into a single representation $\mathbf{H}^L_Q \in \mathbb{R}^{|\mathcal{N}|\times d_\text{GNN}}$. The overall computation process is described as follows:
\begin{align}
\mathbf{E}_Q & = \frac{1}{N}\sum_{i=0}^{N-1} \mathbf{E}_{t_i} \\
\mathbf{H}^L_Q &=  \text{Triple-GNN}(\mathcal{G}, \mathbf{H}_e^0, \mathbf{H}_r^0)
\end{align}
$\mathbf{H}_r^0 \in \mathbb{R}^{|\mathcal{R}|\times d_\text{GNN}}$ denotes the initialized feature representations of relations in $\mathcal{R}$. $\mathbf{H}^0_e \in \mathbb{R}^{|\mathcal{N}|\times d_\text{GNN}}$ represents the initial feature input of entities, for any $\mathbf{h}_{e_i}^0 \in \mathbf{H}_e^0$, its initialization is defined as follows:
\begin{align}
\mathbf{h}_{e_i}^0 = \begin{cases} 
\mathbf{E}_Q, & e_i \in \mathcal{N}_Q, \\
\mathbf{0}, & \text{otherwise}.
\end{cases}
\end{align}

Further details regarding the initialization of the entity and relation inputs ($\mathbf{H}_e^0$ and $\mathbf{H}_r^0$), as well as the subsequent computations within the Triple-GNN, can be found in Appendix \ref{sec:tgnn}. 

After obtaining entity embeddings $\mathbf{H}_Q^L$ via Triple-GNN, we pass them through a linear projection layer, followed by a Sigmoid operation to derive the node probability distribution\footnote{The subscript $\phi_x$
indicates that this module contains trainable parameters proposed in this paper.}:
\begin{align}
\boldsymbol{P}_Q = \mathrm{Sigmoid}(\text{MLP}{\phi_1}(\mathbf{H}_Q^L)) \in \mathbb{R}^{|\mathcal{N}|}
\end{align}

Based on the probability distribution of entities, we select the Top-$\mathcal{K}$ entities with the highest probabilities and construct a candidate entity list \( \mathcal{N}^{'}_Q \):
\begin{align}
\mathcal{N}^{'}_Q = \textbf{\textit{argTop}}\mathcal{K}_{\boldsymbol{p} \in \boldsymbol{P}_Q}(\boldsymbol{p})
\end{align}
where the value of $\mathcal{K}$ is set to $\mid\mathcal{T}\mid*4 $, where $\mid\mathcal{T}\mid$ is the number of triples in \( \mathcal{T} \). 

Upon obtaining the candidate entity list \( \mathcal{N}^{'}_Q \), we anchor each entity $e^{'}_i \in \mathcal{N}^{'}_Q$ within \( \mathcal{G} \) and construct a subgraph connected by their existing relations \( \mathcal{R} \), yielding the final Global Guidance Subgraph denoted as \( \mathcal{G}^{'}_Q \).
The training details of Triple-GNN will be described in Appendix \ref{sec:training}.

\subsection{Structure-Tracing Subgraph Retrieval}
\label{subsec:structure-tracing-subgraph-ret}
In this section, we introduce the Structure-Tracing Subgraph Retrieval module, the primary objective of this module is to identify a specific subgraph within $\mathcal{G}$ that exhibits high structural and semantic isomorphism\footnote{Following the approach of SimGRAG, we say that $P$ and $S$ are isomorphic if there exists a bijective mapping $f:V_P\to V_S$ such that an edge $\langle u, v \rangle$ exists in $P$ if and only if the edge $\langle f(u), f(v) \rangle$ exists in $S$.} \cite{cai2025simgrag} to the query schema graph $\mathcal{G}_{\text{sch}}$. For each question entity $\hat{e} \in \mathcal{N}_Q$, we first retrieve the Top-$N$ ($N=50$) most similar entity nodes $R_{\hat{e}}$ from $\mathcal{G}$, along with their corresponding cosine similarity scores $\boldsymbol{S}_{\hat{e}}$:
\begin{align}
\mathcal{N}_{\hat{e}} &= \textbf{\textit{argTopN}}_{e' \in \mathcal{N}}(\mathrm{Sim}(\hat{e}, e')) \\
\boldsymbol{S}_{\hat{e}} &= \{\mathrm{Sim}(\hat{e}, e') \mid e' \in \mathcal{N}_{\hat{e}}\}
\end{align}

\textbf{Global Structural Consistency Bias.} Since \textbf{Sim} is a shallow entity-to-entity semantic similarity function, we design a global-prior score rectification mechanism by incorporating the structural priors from Guidance Graph \( \mathcal{G}^{'}_Q \). Specifically, for the aforementioned score $\boldsymbol{S}_{\hat{e}}$, the score rectification is applied as follows:
\begin{align}
\boldsymbol{S}^*_{\hat{e}} &= \boldsymbol{S}_{\hat{e}} * \mathbb{I}_\text{Ent}(\mathcal{N}_{\hat{e}}) \nonumber \\
&=\{\mathrm{Sim}(\hat{e}, e')*\mathbb{I}_\text{Ent}(e') \mid e' \in \mathcal{N}_{\hat{e}}\}
\label{entity_bias}
\end{align}
where $\mathbb{I}_\text{Ent}$ is \textbf{Entity-level Global Structural Consistency Bias}, and $\mathbb{I}_\text{Ent}$ is defined as follows:
\begin{align}
\mathbb{I}_\text{Ent}(e) = \begin{cases} 
\boldsymbol{\frac{3}{2}}, & e \in \mathcal{N}^{'}_Q, \\
\boldsymbol{1}, & \text{otherwise}.
\end{cases}
\label{eq:weighting_func}
\end{align}

This means that if an entity $e$ exists in the Global Guidance Subgraph $\mathcal{G}^{'}_Q$, it is considered more important for the query reasoning structure. 

To initiate the subgraph matching search, we first establish a starting anchor mapping between the schema graph \( \mathcal{G}_\text{sch} \) and the knowledge graph \( \mathcal{G} \). Specifically, for the question entity $\hat{e}$, we identify its counterpart node $e^*$ in \( \mathcal{G} \) by selecting the highest-scoring candidate from $\mathcal{N}_{\hat{e}}$ based on $\boldsymbol{S}^*_{\hat{e}}$. Simultaneously, we locate the corresponding node $e$ within \( \mathcal{G}_\text{sch} \) corresponding to the entity
$\hat{e}$ via fuzzy string matching. This establishes the starting pair ($e$, $e^*$). Proceeding from this anchor, we execute the structure-tracing matching: for a specific edge ($e$, $r$, $e'$)\footnote{\textbf{Note:} For brevity and to explicitly delineate the constituent nodes of each edge, we uniformly use ($e$, $r$, $e'$) to denote both the triple $t$ and the edge corresponding to relation $r$.} defined in \( \mathcal{G}_\text{sch} \), we seek a structurally and semantically matching edge ($e^*$, $r^*$, $e^{*'}$) within \( \mathcal{G} \). This matching process is guided by calculating the globally-aware triple score $\mathrm{T\text{-}Score}$ defined as follows:
\begin{align}
\mathrm{T\text{-}Score}((e, r, e'), (e^*, &r^*, e^{*'})) = \nonumber \\
     \quad \mathrm{Sim}(\text{Encoder}(e, r, e'), &\text{Encoder}(e^*, r^*, e^{*'})) \nonumber \\
     \quad + \mathbb{I}_\text{Tri}&((e^*, r^*, e^{*'}))
\label{eq:t-score}
\end{align}
where $\mathrm{Sim}(t_i, t_j)=\frac{\mathbf{E}_{t_i} \cdot \mathbf{E}_{t_j}}{\|\mathbf{E}_{t_i}\|\,\|\mathbf{E}_{t_j}\|}$. Similar to entity nodes retrieval, we incorporate a structural constraint based on the Global Guidance Subgraph \( \mathcal{G}^{'}_Q \) by integrating with a \textbf{Triple-level Global Structural Consistency Bias} $\mathbb{I}_\text{Tri}$. 
The definition of $\mathbb{I}_\text{Tri}$ denotes an Triple-level structural constraint, defined as follows:
\begin{align}
\mathbb{I}_\text{Tri}(t) = \begin{cases} 
\boldsymbol{\frac{1}{2}}, & t \in \mathcal{G}'_Q, \\
\boldsymbol{0}, & \text{otherwise}.
\end{cases}
\label{eq:triple_bias}
\end{align}

The above describes the method for obtaining edge mappings through single-step reasoning using $\mathrm{T\text{-}Score}$. For the entire graph reasoning process, recursive matching is performed until a concrete subgraph $\mathcal{G}_\text{sch}^*$ that is structurally isomorphic to $\mathcal{G}_\text{sch}$. Specifically, at step $i$ of the matching process, the cumulative score being $\mathcal{S}_i$, the score for the next step $\mathcal{S}_{i+1}$ is computed as follows:
\begin{align}
\mathcal{S}_{i+1} &= \mathcal{S}_i + \mathrm{T\text{-}Score}(t_{i+1}, t_{i+1}^*) \\
\mathcal{S}_0 &= \boldsymbol{S}^*_e[e_0^*]
\end{align}

It is important to note that although the KG is directed, the matching process operates in an undirected context. In other words, we do not distinguish between incoming and outgoing edges in the matching selection. We will provide a detailed description of this algorithm in the Appendix \ref{sec:sstr}.
\subsection{Retrieval Behaviors of Different Strategies}
As discussed in Section \ref{subsec:SGDA}, multi-hop questions can be categorized into \textit{Precision} and \textit{Breadth} types. To accommodate these distinct reasoning requirements, STEM adopts an adaptive search strategy that dynamically adjusts the edge selection behavior during the Structure-Tracing Subgraph Retrieval process. Specifically, for \textit{Precision} strategy, we employ a Greedy Selection mechanism by selecting strictly the edge with the maximum score $\mathcal{S}$. For \textit{Breadth} strategy, we employ a Threshold-Based Selection mechanism by retaining all candidate edges whose scores $\mathcal{S}$ exceed a pre-defined confidence threshold $\theta$. In fact, Threshold-Based Selection allows the structure tracing to branch out, transforming the linear search path into a search tree. 

We present detailed illustrations of both selection strategies and experiments to analyze the their impact on performance in the Appendix \ref{sec:appendixd1}. 

\subsection{Generation}
\label{subsec:generation}
Upon completing the subgraph retrieval, we obtain a query-specific evidence subgraph \( \mathcal{G}_\text{reason} \) by integrating the resulting subgraphs from all search processes. To linearize this graph structure into a format compatible with LLM prompting, we perform Depth-First Search starting from the question entity nodes within \( \mathcal{G}_\text{reason} \) to flatten the subgraph into a set of coherent reasoning chains, finally obtain \( \mathcal{C}_\text{reason} \). We apply prompt \hyperref[fig:p3]{\( \mathcal{P}_3 \)} as an LLM instruction and take \( \mathcal{P}_3 \), \( Q \) and \( \mathcal{C}_\text{reason} \) as input, to infer the final answer:
\begin{align}
\mathcal{A} \leftarrow \text{LLM}_{\theta}(\mathcal{P}_3, Q, \mathcal{C}_\text{reason})
\end{align}

\begin{table}[t]
\footnotesize
\centering
\renewcommand{\arraystretch}{1.15} % 行高
\setlength{\tabcolsep}{2.5pt} 

\resizebox{\columnwidth}{!}{
\begin{tabular}{l|cc|cc}
\toprule
\hline
\multirow{2}{*}{\textbf{Model}} & 
\multicolumn{2}{c|}{\textbf{WebQSP}} & 
\multicolumn{2}{c}{\textbf{CWQ}} \\ \cline{2-5} 

& Hit@1 & $F_1$ Score & Hit@1 & $F_1$ Score \\ 
\hline

\multicolumn{5}{c}{\textbf{GPT-4o}} \\ 
\hline
\textbf{GPT-4o}      & 61.8 & 43.6 & 38.2 & 32.9 \\
\textbf{GPT-4o}+Fewshot  & 71.68 & 63.7 & 57.59 & 44.72 \\
\textbf{GPT-4o}+CoT  & 74.12 & 64.25 & 59.36 & 48.24 \\ 
\hline

\multicolumn{5}{c}{\textbf{With Finetuning}} \\ 
\hline
\textbf{NSM}            & 74.31 & - & 53.92 & - \\ \hdashline
\textbf{DeCAF$_{FiD-3B}$}    &  82.1 & - & 70.42 & - \\ \hdashline
\textbf{KD-CoT$_{T5-large}$}           & 73.7 & 50.2 & 50.5 & - \\ \hdashline
\textbf{RoG$_{Llama2-Chat-7B}$} & 83.15 & 69.81 & 61.39 & 56.17 \\
\textbf{RoG$_{Llama-3.1-70B}$} & 86.1 & 68.87 & 67.43 & 60.3 \\ 
\textbf{RoG$_{GPT-4o}$} & 88.09 & 70.12 & 69.61 & 61.97 \\ 
\textbf{LightProf$_{Llama3-8B}$} & 83.8 & - & 59.3 & - \\ \hdashline
\textbf{GRAG$_{LLaMA2-7B}$}         & 72.75 & 50.41 & - & - \\ \hdashline
\textbf{GNN-RAG$_{Llama2-7B}$} & 86.4 & 69.0 & 67.3 & 59.1 \\ 
\hline

\multicolumn{5}{c}{\textbf{With Prompting}} \\ 
\hline
\textbf{Kaping$_{gpt-3.5-turbo}$}             & 72.42 & 65.12 & 53.42 & 50.32 \\ \hdashline
\textbf{ToG$_{gpt-3.5-turbo}$}    & 75.08 & 72.32 & 57.59 & 56.64 \\ \hdashline
\textbf{G-Ret$_{Llama2-7B}$}          & 70.16 & 50.23 & - & - \\ \hdashline
\textbf{PoG$_{gpt-3.5}$}          & 82.0 & - & 63.2 & - \\ \hdashline
\textbf{ReKnoS$_{gpt-3.5}$}      & 81.1 & - & 58.5 & - \\ \hdashline
\textbf{MFC$_{gpt-4o-mini}$} & 78.9 & - & 62.8 & - \\ 
\textbf{SubgraphRAG$_{gpt-4o}$} & 83.1 & - & 56.3 & - \\ 
\textbf{FiDeLiS$_{gpt-4-turbo}$} & 84.39 & \textbf{78.32} & 71.47 & 64.32 \\ \hdashline
\textbf{ProgRAG$_{gpt-4o-mini}$} & 90.4 & - & 73.3 & - \\ 
\hline

\multicolumn{5}{c}{\textbf{Our Proposed Method}} \\ 
\hline
\textbf{STEM$_{Llama-3.1-8B}$}             & 86.63 & 71.05 & 68.76 & 60.81 \\ \hdashline
\textbf{STEM$_{Llama-3.1-70B}$}    & 88.08 & 74.62 & 72.53 & 62.09 \\ \hdashline
\textbf{STEM$_{GPT-4o}$}          & \textbf{90.94} & 76.18 & \textbf{74.09} & \textbf{65.33} \\ \hdashline
\hline
\bottomrule
\end{tabular}
}
\caption{Comparison of different models on WebQSP and CWQ datasets.}
\label{tab:main_results}
\end{table}

\section{Experiments}

In this section, we present the experimental results and analysis. We conduct the experimental process from the following aspects: (1) How does STEM perform on multi-hop reasoning tasks compared to existing classical and state-of-the-art methods? (2) A fine-grained performance analysis across varying answer numbers, reasoning depths, and underlying reasoning models. (3) Ablation studies on the Semantic-to-Structural Projection pipeline and the Global Structural Consistency Bias. Detailed implementation regarding the training of SGDA, SAGB, and Triple-GNN—including data construction processes and training configurations—is provided in Appendix \ref{sec:training}.

Furthermore, we will describe the details of the experimental setup, test datasets, evaluation metrics, baselines, and other experiments and analyses in the Appendix \ref{sec:exp_detail} and Appendix \ref{sec:additional_exp}. Additionally, a systematic analysis of failure modes and error propagation across the STEM pipeline is detailed in Appendix \ref{sec:afmep}.

\subsection{Main Results}
As described in Table \ref{tab:main_results}, the comparative experimental results demonstrates that STEM achieves a significant performance improvement over other models across both datasets. First, compared to methods relying solely on GPT-4o \cite{openai2024gpt4osystemcard}, STEM exhibits an increase of over 10\% in both Hit@1 and F1 Score on both datasets. When compared to fine-tuned models, STEM + Llama-3.1-8B outperforms other baselines with similar parameter scales, including RoG, LightProf and GNN-RAG. The lead is particularly notable on CWQ, where Hit@1 improves by approximately 6\% compared to RoG. Remarkably, it even surpasses the RoG model utilizing the larger Llama-3.1-70B backbone. Meanwhile, STEM + Llama-3.1-70B achieves even greater margins, boosting the F1 score on WebQSP by about 6\% and Hit@1 on CWQ by 5\% compared to RoG + Llama-3.1-70B. When compared with prompting-based methods, STEM + GPT-4o demonstrates a significant advantage over other approaches utilizing the GPT series. Specifically, on WebQSP, it improves Hit@1 by approximately 7\% over the highly competitive baseline, FiDeLiS, although its F1 score is slightly lower. On CWQ, STEM yields distinct improvements across all metrics. Ultimately, STEM + GPT-4o achieves SOTA performance on three out of the four evaluated metrics, with the exception of the F1 score on WebQSP.

\subsection{Performance Analysis}
\label{sec:perform_ana}

We partition the test set based on the number of answers and evaluate performance on each subset, with the results shown in Table \ref{tab:res_answer_count}. It is evident that STEM significantly outperforms both RoG and GNN-RAG across all answer count categories. Notably, STEM achieves an improvement of approximately 4\% on the WebQSP subset with answers
$\geq 10$ and a 9\% gain on the CWQ subset with answers in [2, 4]. These results demonstrate STEM's superior performance in ensuring comprehensive coverage for multi-answer queries.

We stratify the performance by reasoning hop number, with the results presented in Table \ref{tab:res_hop_num}. It is evident that STEM generally maintains a strong competitive edge. However, a notable exception is observed on the CWQ (hop=2), where our method lags significantly behind GNN-RAG.

We conducted a comparative analysis of RoG and STEM using different reasoning models while keeping the retrieval pipeline constant, the results are presented in Table \ref{tab:reason_model}. As the results demonstrate, although replacing the original LLaMA2-Chat-7B used in RoG with Llama-3.1-70B-Ins or GPT-4o yields performance gains, STEM maintains a competitive edge under identical reasoning model configurations. 

However, we acknowledge a potential ambiguity in this analysis: the performance improvements might stem partially from the superior parametric knowledge of these advanced models, rather than solely from enhancements in reasoning capability. To investigate this further, we conducted an additional experiment to assess the coverage rate of evidence subgraph retrieval. Specifically, we calculated the proportion of the ground-truth reasoning path covered by the evidence subgraph obtained via Structure-Tracing Subgraph Retrieval. Specifically, given a question $Q$, let $\mathcal{R}$ denote the ground-truth reasoning path of $Q$, and $\mathcal{G}_\text{reason}$ denote the evidence subgraph. Coverage rate is defined as:
\begin{align}
\text{Coverage\_rate} = \frac{\mid \mathcal{R} \cap\mathcal{G}_\text{reason}\mid}{\mid \mathcal{R} \mid}
\label{coverage}
\end{align}

The results are presented in Table \ref{tab:path_cov}, indicating that the coverage rate of the evidence subgraph gradually decreases as the number of answers increases, yet it remains at a relatively high level. Furthermore, the coverage on CWQ is consistently lower than on WebQSP due to the question complexity.

\begin{table}[t]
\centering
\scriptsize
\renewcommand{\arraystretch}{1.15} % 行高
\setlength{\tabcolsep}{2.5pt} 

\resizebox{\columnwidth}{!}{
\begin{tabular}{c|c|cccc}

\hline
{\textbf{Method}} & {\textbf{Dataset}} & {\textbf{Ans} = 1} & {\textbf{Ans} $\in$ [2,4]} & {\textbf{Ans} $\in$ [5,9]} & {\textbf{Ans} $\geq$ 10} \\
\hline
\multirow{2}{*}\textbf{RoG} & \textbf{WebQSP} & 67.89 & 79.39 & 75.04 & 58.33 \\

& \textbf{CWQ} & 56.90 & 53.73 & 58.36 & 43.62 \\ 

\hline
\multirow{2}{*}\textbf{GNN-RAG} & \textbf{WebQSP} & 71.24 & 76.30 & 74.06 & 56.28 \\

& \textbf{CWQ} & 60.40 & 55.52 & 61.49 & 50.08 \\ 

\hline
\multirow{2}{*}\textbf{STEM+GPT-4o} & \textbf{WebQSP} & 75.26 & 81.87 & 78.38 & 62.46 \\

& \textbf{CWQ} & 65.32 & 64.35 & 66.37 & 53.86 \\ 
\hline
\end{tabular}
}
\caption{Detailed results (F1) grouped by the number of answers.}
\label{tab:res_answer_count}
\end{table}

\begin{table}[t]
\centering
\scriptsize
\renewcommand{\arraystretch}{1.15} % 行高
\setlength{\tabcolsep}{2.5pt} 

\begin{tabular}{c|c|ccc}

\hline
{\textbf{Method}} & {\textbf{Dataset}} & {\textbf{Hop} = 1} & {\textbf{Hop} = 2} & {\textbf{Hop} $\geq 3$} \\
\hline
\multirow{2}{*}\textbf{RoG} & \textbf{WebQSP} & 77.03 & 64.86 & -  \\

& \textbf{CWQ} & 62.88 & 58.46 & 37.82  \\ 

\hline
\multirow{2}{*}\textbf{GNN-RAG} & \textbf{WebQSP} & 72.0 &  69.8 & -  \\

& \textbf{CWQ} & 47.4 & 69.4 & 51.8  \\ 

\hline
\multirow{2}{*}\textbf{STEM+GPT-4o} & \textbf{WebQSP} & 81.49 & 75.35 & - \\

& \textbf{CWQ} & 67.46 & 66.73  & 52.15 \\ 
\hline
\end{tabular}
\caption{Detailed results (F1) grouped by the maximum reasoning hop number.}
\label{tab:res_hop_num}
\end{table}

\begin{table}[t]
\centering
\scriptsize
\renewcommand{\arraystretch}{1.15} % 行高
\setlength{\tabcolsep}{2.5pt} 

\begin{tabular}{c|c|cc|cc}

\hline
\multirow{2}{*}{\textbf{Method}} & \multirow{2}{*}{\textbf{Reasoning Model}} & 
\multicolumn{2}{c|}{\textbf{WebQSP}} & 
\multicolumn{2}{c}{\textbf{CWQ}} \\ \cline{3-6}

 &  &  Hit@1 & $F_1$ Score & Hit@1 & $F_1$ Score \\ 
\hline
\multirow{3}{*}{RoG} & LLaMA2-Chat-7B & 83.15 & 69.81 & 61.39 & 56.17 \\
& Llama-3.1-70B-Ins & 86.1 & 68.87 & 67.43 & 60.3 \\ 
& GPT-4o & 88.09 & 70.12 & 69.61 & 61.97 \\ 
\hline
\multirow{2}{*}{STEM} & Llama-3.1-70B-Ins & 88.08 & 74.62 & 72.53 & 62.09 \\ 
& GPT-4o & 90.94 & 76.18 & 74.09 & 65.33 \\ 

\hline
\end{tabular}
\caption{Impact of reasoning models on performance.}
\label{tab:reason_model}
\end{table}

\begin{table}[t]
\centering
\scriptsize
\renewcommand{\arraystretch}{1.15} % 行高
\setlength{\tabcolsep}{2.5pt} 

\begin{tabular}{c|cccc}

\hline
{\textbf{Dataset}} & {\textbf{Ans} = 1} & {\textbf{Ans} $\in$ [2,4]} & {\textbf{Ans} $\in$ [5,9]} & {\textbf{Ans} $\geq$ 10} \\
\hline
\textbf{WebQSP} & 81.9 & 76.64 & 71.45 & 58.23 \\ 
\hline
\textbf{CWQ} & 74.28 & 66.57 & 62.71 & 51.89 \\
\hline
\end{tabular}
\caption{Coverage Rate (\%) of ground-truth reasoning paths within evidence subgraphs.}
\label{tab:path_cov}
\end{table}

\subsection{Ablation study}
\textbf{Semantic-to-Structural Projection Pipeline.} We conduct ablation studies by comparing our proposed Semantic-to-Structural Projection pipeline against powerful off-the-shelf LLMs to evaluate its effectiveness\footnote{For baseline LLMs, we implement the pipeline using a few-shot prompting approach, adopting the identical prompt templates employed by the SGDA and SAGB. Regarding the answer strategy, if a valid strategy cannot be successfully extracted after 5 retries, we default to \textit{Precision} for the current query.}. The comparative results are presented in Table \ref{tab:pipline}. Our method demonstrates a significant performance advantage over the baseline models, surpassing the strongest competitor by over 23\% on the CWQ dataset. Among the baselines, GPT-4o consistently outperforms Llama-3.1-70B-Ins, reflecting its superior few-shot KG alignment. Our results validate the critical role of logic-aware projection in KG reasoning. 

\begin{table}[t]
\footnotesize
\centering
\renewcommand{\arraystretch}{1.15} % 行高
\setlength{\tabcolsep}{2pt} 

\resizebox{\columnwidth}{!}{
\begin{tabular}{l|cc|cc}
\toprule
\hline
\multirow{2}{*}{\textbf{Pipelines}} & 
\multicolumn{2}{c|}{\textbf{WebQSP}} & 
\multicolumn{2}{c}{\textbf{CWQ}} \\ \cline{2-5} 

& Hit@1 & $F_1$ Score & Hit@1 & $F_1$ Score \\ 

\hline
\textbf{Llama-3.1-70B-Ins}    & 77.74 & 61.21 & 46.68 & 41.83  \\
\textbf{GPT-4o}  & 83.14 & 65.77 & 50.43 & 43.2 \\
\textbf{Our Pipeline}  & 90.94 & 76.18 & 74.09 & 65.33 \\ 
\hline
\bottomrule
\end{tabular}
}
\caption{Comparison of different Question Planing Pipelines.}
\label{tab:pipline}
\end{table}

\textbf{Global Structural Consistency Bias.} As the Global Guidance Subgraph serves as a critical structural prior, we conducted an ablation study by selectively removing its bias terms. We evaluated three variants: \textbf{w/o Entity Bias} (removing the calculation of the indicator term \( \mathbb{I}_\text{Ent} \) in Equation \ref{entity_bias}), \textbf{w/o Triple Bias} (removing the calculation of \( \mathbb{I}_\text{Tri} \) from Equation \ref{eq:t-score}), and \textbf{w/o Both}. 

The ablation results are presented in Table \ref{tab:bias_abla}. We observe that incorporating both Entity-level and Triple-level Biases yields significant performance gains. Conversely, removing Triple-level Bias leads to a marked decline, particularly in the Hit@1 metric on WebQSP and across all metrics on CWQ, with performance drops reaching up to 10\% on CWQ. Removing both biases causes further degradation, with the Hit@1 score on CWQ dropping by an additional 3\% compared to the w/o Entity Bias setting, while the F1 score on WebQSP plummets by nearly 5\%. Furthermore, the w/o Triple-level setting yields inferior performance compared to the w/o Entity-level, indicating that the Triple-level bias plays a more critical role. These results revealing the limitations of relying solely on local semantic matching. Appendix \ref{sec:ggca}  details its error-correction analysis.

We conducted a comprehensive set of additional experiments that are equally critical to validating the robustness of STEM; however, due to space constraints, these results are detailed in Appendix \ref{sec:additional_exp}. The supplementary evaluations encompass fine-grained performance analyses and sensitivity studies on key hyperparameters, including the Answer Strategy $\sigma$, Initial Entity Count $\mathcal{K}$, and SGDA Beam Size $\mathcal{B}$, as well as parameter tuning for the Global Structural Consistency Bias \( \mathbb{I}_\text{Ent} \) and \( \mathbb{I}_\text{Tri} \). Furthermore, we provide in-depth Case Studies, Efficiency Analysis, and Interpretability Analysis.

\begin{table}[t]
\footnotesize
\centering
\renewcommand{\arraystretch}{1.15}
\setlength{\tabcolsep}{2.5pt} 

\begin{tabular}{l|cc|cc}
\toprule
\hline
\multirow{2}{*}{\textbf{Scoring Bias}} & 
\multicolumn{2}{c|}{\textbf{WebQSP}} & 
\multicolumn{2}{c}{\textbf{CWQ}} \\ \cline{2-5} 

& Hit@1 & $F_1$ Score & Hit@1 & $F_1$ Score \\ 

\hline
\textbf{STEM$_{GPT-4o}$}  & 90.94 & 76.18 & 74.09 & 65.33 \\
\textbf{\;\;\;\; w/o $\mathbb{I}_\text{Ent}$ $\&$ $\mathbb{I}_\text{Tri}$}  & 86.31 & 70.80 & 63.91 & 55.59 \\
\textbf{\;\;\;\; w/o $\mathbb{I}_\text{Ent}$}    & 86.45 & 75.81 & 66.35 & 57.35  \\
\textbf{\;\;\;\; w/o $\mathbb{I}_\text{Tri}$} & 86.95 & 73.45 & 64.90 & 56.42  \\ 
\hline
\bottomrule
\end{tabular}
\caption{Ablation Study on Entity-level and Triple-level Scoring Biases.}
\label{tab:bias_abla}
\end{table}

\section{Related Work}
\subsection{Retrieval-Augmented Generation (RAG)}
Retrieval-Augmented Generation (RAG) has emerged as a dominant paradigm to mitigate the hallucination issues of Large Language Models (LLMs) \citep{lewis2020retrieval, guu2020realm, karpukhin2020dense, izacard2021unsupervised, asai2023selfrag}. Adaptive-RAG \citep{jeong2024adaptiverag} dynamically selects retrieval strategies based on query complexity. Similarly, Corrective RAG (CRAG) \citep{yan2024corrective} incorporates a lightweight evaluator to filter irrelevant documents and trigger web searches as a fallback. 
\subsection{Multi-hop Reasoning in RAG}
For multi-step deduction, research has evolved from single-step retrieval to Iterative and Chain-of-Thought (CoT) Retrieval \citep{trivedi2022interleaving}. ReAct \citep{yao2022react} further generalizes this by modeling LLMs as agents that can perform search actions. Chain-of-Note \citep{yu2023chainofnote} generates sequential reading notes to evaluate document relevance before aggregation. Demonstrate-Search-Predict (DSP) \citep{khattab2022demonstrate} uses frozen LMs to orchestrate sophisticated retrieval pipelines via natural language programs. Tree of Clarifications \citep{kim2023tree} constructs a tree of disambiguations to handle ambiguous questions recursively. 
\subsection{Knowledge Graph-based RAG}
Knowledge Graphs offer a promising solution to the reasoning disconnection problem to encode graph structures \citep{yasunaga2021qa, zhang2022greaselm}, but often struggled to scale or integrate flexibly with LLMs. GraphRAG \citep{edge2024graphrag} introduces a community-detection-based approach to generate hierarchical summaries of the graph for global query understanding. For multi-hop reasoning, GNN-RAG \citep{mavromatis-karypis-2025-gnn} combines GNN-based retrieval with LLM reasoning to handle complex graph topology. Other works like StructGPT \citep{jiang2023structgpt} and KAPING \citep{baek-etal-2023-knowledge} explore zero-shot prompting strategies to interface LLMs with structured data. 

\section{Conclusion}
We presented Structure-Tracing Evidence Mining, a novel framework that shifts multi-hop KG-RAG from sequential path finding to holistic subgraph matching. By synergizing a fine-tuned Semantic-to-Structural Projection pipeline with a Triple-Dependent GNN, STEM effectively bridges the gap between natural language and KG schemas, retrieving logically connected evidence subgraphs and align with the knowledge structure. Extensive experiments on WebQSP and CWQ demonstrate that STEM significantly outperforms existing baselines, achieving new State-of-the-Art results. 

\section*{Limitations}
Although STEM effectively bridges the semantic-structural gap, challenges persist due to the inherent diversity of KG topologies. First, in highly complex reasoning tasks, planning deviations may still occur, leading to scenarios where all generated candidate schema graphs fail to match the factual structure, thereby resulting in retrieval failure. Second, STEM relies on domain-specific fine-tuning and access to the target KG's structure. While achieving strong performance on Freebase-based benchmarks (WebQSP, CWQ), we acknowledge it is not a general-purpose zero-shot method; this dependency limits its transferability to unseen KGs or novel schema types. Finally, regarding efficiency, the threshold-based expansion in the \textit{Breadth} strategy increases computational latency, we consider this a necessary trade-off for answer exhaustiveness. Moreover, this expansion is selective—occurring not at every step, but exclusively when the retrieval of multiple simultaneous answer paths is required. 

\section*{Ethical Considerations}

We address the ethical considerations and potential risks as follows:

\paragraph{Data Provenance and Licensing.}
The knowledge graph utilized in this study, is a widely adopted, publicly available database distributed under the CC-BY license. Our usage of WebQSP and CWQ datasets strictly adheres to their respective data usage policies and licenses. These datasets are standard benchmarks in the research community and do not contain private, personally identifiable information (PII), or offensive content that would require special redaction for this study.

\paragraph{Bias and Fairness.}
We acknowledge that KGQA systems are susceptible to propagating biases inherent in their underlying knowledge sources. Specifically, the Freebase KG is known to exhibit significant demographic, cultural, and geographical skews. Since STEM is designed to be faithful to the retrieved subgraph, it inevitably reflects the distributional properties of the source KG. Therefore, users should interpret the outputs of STEM as a reflection of the facts stored in the specific Knowledge Graph, rather than as an unbiased representation of real-world truth. 

\section*{Acknowledgments}
We would like to thank the Action Editors and the anonymous reviewers for their constructive feedback and insightful comments, which helped improve the quality of this paper.

\bibliography{custom}
\newpage
\appendix

\section{Triple-GNN Implementation}
\label{sec:tgnn}

In this section, we describe the overall execution process of the Triple-GNN. Specifically, we first employ an Pretrained Embedding Model (PEM) to obtain vector representations for both the triples and the entities:
\begin{align}
    \mathbf{E}_x &= \text{Encoder}(x) \in \mathbb{R}^{d_{\text{PEM}}} \\
    \mathbf{E}_t &= \text{MLP}_{\phi_2}([\mathbf{E}_{e}; \mathbf{E}_{r}; \mathbf{E}_{e'}]) \in \mathbb{R}^{d_{\text{GNN}}}
\end{align}
for all \(x \in \{e, r, e'\}\), where the terms $e$, $r$ and $e'$ represent the head entity, relation, and tail entity of the triple $t$, respectively. The [;] operation denotes the concatenation of the embedding vectors, transforming the dimensionality from $\mathbb{R}^{d_\text{PEM}} \rightarrow \mathbb{R}^{3d_\text{PEM}}$. 

Subsequently, the MLP operation refers to a linear transformation that maps the concatenated vector from $\mathbb{R}^{3d_{\text{PEM}}}\rightarrow\mathbb{R}^{d_\text{GNN}}$. We finally obtain the embedding representation $\mathbf{E}_t$ of the triple $t$.

Given the structured triples \( \mathcal{T} = \{ t_1, t_2, ... , t_N \} \) obtained from the Semantic-to-Structural Projection pipeline based on query $Q$, with the corresponding knowledge graph $\mathcal{G}$ , entity set $\mathcal{N}$ and relation set 
$\mathcal{R}$, we first compute the embedding for each triple to yield \( \mathbf{E}_\mathcal{T} = \{ \mathbf{E}_{t_1}, \mathbf{E}_{t_2}, ... , \mathbf{E}_{t_N} \} \), then aggregate the embedding set into a single representation $\mathbf{E}_Q \in \mathbb{R}^{d_\text{GNN}}$ via average pooling, and finally process it through the $L$-layer Triple-GNN to obtain $\mathbf{H}^L_Q \in \mathbb{R}^{|\mathcal{N}|\times{d_\text{GNN}}}$:
\begin{align}
\mathbf{E}_Q &= \frac{1}{N}\sum_{i=1}^{N} \mathbf{E}_{t_i} \\
\mathbf{H}^L_Q &= \text{Triple-GNN}_{\phi_3}(\mathcal{G}, \mathbf{H}^0_e, \mathbf{H}^0_r)
\end{align}
$\mathbf{H}^0_e$ represents the initial feature input of entities and $\mathbf{H}^0_e \in \mathbb{R}^{|\mathcal{N}|\times{d_\text{GNN}}}$, for any $\mathbf{h}_{e_i}^0 \in \mathbf{H}^0_e$, its initialization method is defined as follows:
\begin{align}
\mathbf{h}_{e_i}^0 = \begin{cases} 
\mathbf{E}_Q, & e_i \in \mathcal{N}_{Q}, \\
\mathbf{0}, & \text{otherwise}.
\end{cases}
\end{align}
where $\mathcal{N}_{Q}$ denotes the set of question entities for question $Q$.

$\mathbf{H}^0_r \in \mathbb{R}^{|\mathcal{R}|\times d_\text{GNN}}$ denotes the initialized feature representations of relations in $\mathcal{R}$. For each $\mathbf{h}^0_{r_i} \in \mathbf{H}^0_r$, we initialize it using the same Encoder employed for the triple embeddings, followed by an MLP for linear projection, as defined below:
\begin{align}
    \mathbf{z}_{r_i} &= \operatorname{Encoder}(r_i) \in \mathbb{R}^{d_{\text{PEM}}}, \label{eq:rel_enc} \\
    \mathbf{h}^0_{r_i} &= \operatorname{MLP}_{\phi_4}(\mathbf{z}_{r_i}) \in \mathbb{R}^{d_{\text{GNN}}}, \quad \forall r_i \in \mathcal{R} \label{eq:rel_proj}
\end{align}
the MLP projects the encoded relation representation from 
$\mathbb{R}^{d_\text{PEM}}$ into the $\mathbb{R}^{d_\text{GNN}}$ space.

We now describe the message passing computational flow of a single layer of Triple-GNN. Building upon the Query-Dependent GNN design proposed by GFM-RAG \cite{luo2025gfmrag_arxiv}, for the embedding representation $\mathbf{h}_e^{L-1}$ of node $e$ obtained at the ($L$-1)-th GNN layer, the representation at the $L$-th layer is computed as follows:
\begin{align}
\mathbf{m}_e^{L}& = \text{MSG}(\mathbf{h}_e^{L-1}, g^{L}(\mathbf{h}_r^{L-1}), \mathbf{h}_{e'}^{L-1}), \\
\mathbf{h}_e^{L}& = \text{Update}(\mathbf{h}_e^{L-1}, \text{Agg}(\mathcal{M}_{e}^L))
\end{align}
where $(e, r, e') \in \mathcal{G}$, and $\mathcal{M}_{e}^L =\{ \mathbf{m}^L_{e'} | e' \in \text{N}_r(e), r \in \mathcal{R} \}$. The set $\text{N}_r(e)$ denotes the collection of all neighboring nodes of entity $e$ under relation $r \in \mathcal{R}$. 

In the context of a triple $(e, r, e')$ associated with the entity $e$, the notations $\mathbf{h}_e^{L-1}$, $\mathbf{h}_r^{L-1}$, and $\mathbf{h}_{e'}^{L-1}$ represent the embedding representations of the head entity $e$, the relation $r$, and the tail entity $e'$ at layer $L$-1, respectively. The operation denoted as \text{MSG} employs a \text{DistMult} \cite{yang2015embedding} function to process the triple. The function $g^L$ constitutes a 2-layer MLP operation at the subsequent layer $L$. The \text{Agg} operation collects the states $\mathbf{m}_{e'}^L$ of all neighbor nodes $e'$ of $e$ and performs a mean reduction:
\begin{align}
\text{Agg}(\mathcal{M}_{e}^L) = \frac{1}{\mid \mathcal{M}_{e}^L \mid}\sum_{\textbf{m}_{e'}^L \in \mathcal{M}_{e}^L} \textbf{m}_{e'}^L
\end{align}

The \text{Update} function performs the node update operation, achieved by fusing the aggregated neighbor features $\mathcal{M}_{e}^L$ into the current node $\textbf{h}_e^{L-1}$. Specifically, the expression for Update is defined as follows:
\begin{align}
\text{Update}(\textbf{h}_e^{L-1}, \text{Agg}(\mathcal{M}_{e}^L)) = \nonumber \\
\text{MLP}([\textbf{h}_e^{L-1}; \text{Agg}(\mathcal{M}_{e}^L)])
\end{align}
the MLP is designed to map the concatenated 
$\mathbb{R}^{2{d_\text{GNN}}}$ intermediate state back to an $\mathbb{R}^{d_\text{GNN}}$ representation, acting as an effective fusing mechanism of the neighbor features.

\section{Experiment Details}
\label{sec:exp_detail}
\subsection{Datasets and Base KG}
We conduct experiments on two publicly available datasets for multi-hop reasoning, including WebQuestionsSP (WebQSP) \cite{yih-etal-2016-value} and ComplexWebQuestions (CWQ) \cite{talmor-berant-2018-web}. WebQSP is a large-scale multi-hop question-answering dataset, which comes with a knowledge graph in the Freebase\footnote{\url{https://github.com/microsoft/FastRDFStore}} format. CWQ is a more difficult and challenging version of such datasets. Specifically, we utilized the open-source data format provided by RoG \cite{luo2024reasoning}\footnote{\url{https://huggingface.co/datasets/rmanluo/RoG-webqsp}}\footnote{\url{https://huggingface.co/datasets/rmanluo/RoG-cwq}}, as it contains complete queries paired with their corresponding subgraphs extracted from Freebase. To ensure fairness and consistency across experiments, we partitioned all datasets according to RoG, obtaining separate training and test splits. We present the statistics for all datasets in the Table \ref{tab:data_stat}. The distribution of answer counts in the dataset is presented in Table \ref{tab:ans_stat}. 

\begin{table*}[t]
\centering
\renewcommand{\arraystretch}{1.15} % 行高
\setlength{\tabcolsep}{5pt} 

\begin{tabular}{c|ccccccc}
\hline
{\textbf{Dataset}} & {\textbf{Train}} & {\textbf{Dev}} & {\textbf{Test}} & \textbf{Hops} & \textbf{1 Hop} & \textbf{2 Hops} & \textbf{$\geq$3 Hops} \\

\hline
\textbf{WebQSP}    & 2,826 & 239 & 1,628 & \{1,2\} & 65.5\% & 34.5\% & - \\
\textbf{CWQ}  & 27,639 & 3297 & 3,531 & \{1,2,3,4\} & 41\% & 38.3\% & 20.7\% \\
\hline
\end{tabular}
\caption{Statistics of the number of WebQSP and CWQ dataset splits along with the question hops.}
\label{tab:data_stat}
\end{table*}

\begin{table*}[t]
\centering
\renewcommand{\arraystretch}{1.15} % 行高
\setlength{\tabcolsep}{5pt} 

\begin{tabular}{c|cccc}
\hline
{\textbf{Dataset}} & {\textbf{Ans} = 1} & {\textbf{Ans} $\in$ [2,4]} & {\textbf{Ans} $\in$ [5,9]} & {\textbf{Ans} $\geq$ 10} \\

\hline
\textbf{WebQSP}    &  51.8\% & 27.1\% & 8.1\% & 13.0\% \\
\textbf{CWQ}  & 71.4\% & 19.0\% & 5.9\% & 3.7\% \\
\hline
\end{tabular}
\caption{Distribution statistics of answer counts in the two datasets.}
\label{tab:ans_stat}
\end{table*}

\subsection{Implementation Details}
STEM involves three LLM-based modules: SGDA, SAGB, and the LLM reasoning model. For the first two modules, we fine-tune Qwen3-8B\footnote{\url{https://huggingface.co/Qwen/Qwen3-8B}} respectively, and for reasoning model, we select Llama-3.1-8B-Instruct\footnote{\url{https://huggingface.co/meta-llama/Llama-3.1-8B-Instruct}}, Llama-3.1-70B-Instruct\footnote{\url{https://huggingface.co/meta-llama/Llama-3.1-70B-Instruct}}, and GPT-4o\footnote{\url{https://chatgpt.com/}} \cite{openai2024gpt4osystemcard}. To ensure experimental reproducibility, we set the temperature of the reasoning model to 0. Additionally, we configured the SGDA with a beam size \( \mathcal{B} \) of 4 and the SAGB with a temperature of 0. For feature embedding, we use the Qwen3-Embedding-0.6B\footnote{\url{https://huggingface.co/Qwen/Qwen3-Embedding-0.6B}} \cite{zhang2025qwen3embedding} model as the pretrained embedding model (i.e. $d_\text{PEM}$=1024), the Triple-GNN is configured with $L$=6 layers, other configurations are consistent with the settings described in GFM-RAG \cite{luo2025gfmrag_arxiv}(i.e. $d_\text{GNN}$=512). To determine the threshold $\theta$ for the threshold-based search employed in the \textit{Breadth} strategy, we conducted a parameter search on the validation sets of WebQSP and CWQ, ultimately setting $\theta
$=0.6. We performed three independent runs of the full experimental pipeline—encompassing both module training and retrieval-inference testing—and report the average values across all metrics. All models employed in this study were used in strict accordance with their respective licenses and terms of use: OpenAI Terms of Use for GPT-4o, Llama 3.1 Community License for Llama models, and Apache 2.0 License for Qwen models.
\subsection{Evaluation Metrics}
Following established evaluation protocols, we assess model performance using two standard metrics: \textbf{Hits@1} and \textbf{F1 score}. Hits@1 quantifies the accuracy of the top-ranked answer prediction, while the F1 score provides a balanced assessment of answer coverage.
\subsection{Baselines}
Our experimental baselines are categorized into four groups: \textbf{Pure LLM Reasoning}, referring to methods that utilize only the large language model's inherent capabilities, \textbf{With Finetuning}, referring to methods that involve fine-tuning the reasoning model, \textbf{With Prompting}\footnote{Notably, comparing STEM against prompting baselines does not imply zero-shot parity. Rather, it demonstrates that when training is feasible, structural alignment provides substantial gains in factual precision and hallucination mitigation. }, referring to methods that control the reasoning and answering behavior of large language models through prompting, and \textbf{Our Proposed Method}. Notably, our proposed STEM inherently belongs to the fine-tuning category, as it relies on fine-tuned upstream modules for retrieval. We now introduce each category as follows: 

\textbf{Pure LLM Reasoning.}
We do not employ any retrieval components, but instead rely solely on the inherent reasoning capabilities, the model selected is GPT-4o, and experiments are conducted using three distinct settings: pure reasoning, few-shot learning, and CoT prompting.

\textbf{With Finetuning.}
We select the following methods for comparison: \textbf{NSM} \cite{he2021improving} proposes a teacher network to supervise the intermediate reasoning process. \textbf{DeCAF} \cite{yu2023decaf} performs question-relevant retrieval at the document level and constructs logic forms for answers to optimize responses. \textbf{KD-CoT} \cite{wang2023knowledgedriven} proposes KG-guided intermediate reasoning verification to ensure a more reliable reasoning process. \textbf{RoG} \cite{luo2024reasoning} leverages LLM-generated reasoning paths and continuously refines them with KG before performing retrieval. \textbf{GRAG} \cite{hu-etal-2025-grag} optimizes subgraph retrieval complexity and employs both text view and graph view to enhance question comprehension, and \textbf{LightProf} \cite{ao2025lightprof} retrieves the reasoning path, then integrate KG factual and structural information into embeddings for improved answering.

\textbf{With Prompting.}
We adopt the following approaches as baselines for comparison: \textbf{G-Ret} (G-Retriever) \cite{he2024gretriever} proposes a novel RAG framework that formulates subgraph retrieval as a Prize-Collecting Steiner Tree (PCST) problem. \textbf{ToG} \cite{sun2024thinkongraph} introduces a framework where an LLM acts as an agent to iteratively explore reasoning paths on a knowledge graph via beam search. \textbf{Kaping} \cite{baek-etal-2023-knowledge} is a zero-shot framework that retrieves relevant facts from a knowledge graph and prepends them to the input prompt. \textbf{FiDeLiS} \cite{sui-etal-2025-fidelis} proposes a training-free framework that combines step-wise beam search with a deductive scoring function and Path-RAG module. \textbf{PoG} \cite{chen2024planongraph} decomposes questions into sub-objectives and iteratively adapts reasoning paths through guidance, memory, and reflection mechanisms. \textbf{ReKnoS} \cite{sun2024think} introduces a novel framework that enhances LLM reasoning by incorporating super-relations in knowledge graphs. \textbf{MFC} \cite{zhang2025good} transforms questions into knowledge graph triples using LLMs and quantifies question quality based on cognitive metrics. \textbf{SubgraphRAG} \cite{li2024simple} decouples the roles of knowledge graphs and LLMs in RAG systems. \textbf{GNN-RAG} \cite{mavromatis-karypis-2025-gnn} leverages lightweight GNNs for efficient graph retrieval. \textbf{ProgRAG} \cite{park2025prograg} introduces feedback-aware and evidence-aware mechanisms to progressively align LLM reasoning with factual knowledge from graphs.

\section{Training Setup}
\label{sec:training}
\subsection{Basic Training Configuration}
\label{sec:basic-training}
Our work involves the training of three modules: Schema-Grounded Decomposition Agent, Symbol-Aligned Graph Builder, and Triple-GNN\footnote{Here, the Triple-GNN encompasses not only the parameters of the GNN module itself but also the parameters of various associated projection layers: 
$\mathbf{MLP}_{\phi_1}$, $\mathbf{MLP}_{\phi_2}$ and $\mathbf{MLP}_{\phi_4}$, which have been denoted in their respective formula descriptions.}. We will sequentially introduce the data construction and training methods for each of these modules. 

First, we introduce the training data construction method, we utilize the training splits of the WebQSP and CWQ datasets. For both datasets, we take a question $Q$ from the training split, along with its corresponding answer $\mathcal{A}$, question entities set $\mathcal{N}_Q$ and KG $\mathcal{G}$. We first extract reasoning chain $\mathcal{R}$ from $\mathcal{G}$ employing the method proposed in CoG \cite{zhao-etal-2025-correcting}, then decompose the reasoning chain into individual triples $\mathcal{T}$:
\begin{align}
\mathcal{T}=\{t_1, t_2, ..., t_N\}
\end{align}

Then, based on the question $Q$, we mark all entities in $\mathcal{T}$ that do not appear in $\mathcal{N}_Q$ using placeholders, this is because these entities are not question entities, but rather intermediate answer entities in the multi-hop reasoning process. This marking complies with the following principles: (1) when two triples share the same answer entity (indicating that they are connected in the graph), the same identifier ``[ENTX]'' is used; (2) different entities are distinguished by different identifiers (``[ENTX]'' and ``[ENTY]''). After formatting process, we obtain a new masked $T$:
\begin{align}
\mathcal{T}^{'}=\{t^{'}_1, t^{'}_2, ..., t^{'}_N\}
\end{align}

Based on the obtained $\mathcal{T}^{'}$, we generate an assertion for each $t^{'} \in \mathcal{T}^{'}$ by using a prompt \hyperref[fig:p4]{\( \mathcal{P}_4 \)} to instruct a large language model (for all prompt-based data generation tasks in this study, we consistently utilized \textbf{Gemini 2.5 Pro} \cite{comanici2025gemini} API, with the temperature set to 1.0), thereby obtaining a set $\mathcal{S}$ containing assertions corresponding to each triple in $\mathcal{T}^{'}$:
\begin{align}
\mathcal{S}&\leftarrow \text{LLM}(\mathcal{P}_4, \mathcal{T}^{'}, Q)
\end{align}

For \textbf{Symbol-Aligned Graph Builder Training}, we treat the generated assertions $\mathcal{S}_j$ as the source input and the original structured triples $\mathcal{T}^{'}_j$ as the target output to optimize the SAGB model:
\begin{equation}
    \mathcal{L}_{\text{SAGB}} = - \sum_{j=1}^{|\mathcal{D}|} \sum_{k=1}^{|\mathcal{T}^{'}_j|} \log P_{\phi_5}(y_{j,k} \mid \mathcal{P}_2, \mathcal{S}_j, y_{j,<k})
\end{equation}
where $\mathcal{D}$ represents the training dataset, \hyperref[fig:p2]{$ \mathcal{P}_2 $} denotes the instruction prompt used by the SAGB for schema graph building as introduced in Section \ref{subsec:SAGB}.

For question-answering strategy generation, we determine based on the number of ground‑truth answers for $Q$: if there is a single answer, $\sigma$ is set to ``Precision''; if there are multiple answers, $\sigma$ is set to ``Breadth''. Thus for \textbf{Schema-Grounded Decomposition Agent} training, we use $Q_j$ as the source
input and ($\mathcal{S}_j$, $\sigma_j$) as the target output to optimize the SGDA model:
\begin{align}
    \mathcal{L}_{\text{SGDA}}&=  \nonumber\\ - \sum_{j=1}^{|\mathcal{D}|} \sum_{k=1}^{|(\textit{S}_j, \sigma_j)|} \log P_{\phi_6}&(y_{j,k} \mid \mathcal{P}_1, Q_j, y_{j,<k})
\end{align}

For Triple-GNN Training, we apply the method described in Appendix \ref{sec:tgnn} to obtain the pooled embedding representation $\mathbf{E}_Q \in \mathbb{R}^{d_\text{GNN}}$ of $\mathcal{T}^{'}$, then fed into the \textit{L}-layer Triple-GNN to produce $\mathbf{H}_Q$, and finally transformed via a mapping function and activated to obtain the entity probability:
\begin{align}
\mathbf{H}_Q^L = & \text{Triple-GNN}_{\phi_3}(\mathcal{G}, \mathbf{H}_e^0, \mathbf{H}_r^0) \\
\boldsymbol{P}_Q = & \mathrm{Sigmoid}(\text{MLP}_{\phi_1}(\mathbf{H}_Q^L))
\end{align}

The definition of $\mathbf{H}_e^0 \in \mathbb{R}^{|\mathcal{N}|\times{d_\text{GNN}}}$ and $\mathbf{H}_r^0 \in \mathbb{R}^{|\mathcal{R}|\times{d_\text{GNN}}}$ is the same as in Appendix \ref{sec:tgnn}, $\mathcal{N}$ and $\mathcal{R}$ denote the set of entity nodes and relations in graph $\mathcal{G}$ respectively. For the label of each entity $e_i \in \mathcal{N}$, we construct it as follows:
\begin{align}
\boldsymbol{y}_{e_i} = \begin{cases} 
\boldsymbol{1}, & e_i \in \mathcal{T}, \\
\boldsymbol{0}, & \text{otherwise}.
\end{cases}
\end{align}

The final training loss employs the BCE loss \cite{luo2025gfmrag_arxiv}, and is formulated as follows:
\begin{align}
\mathcal{L}_\text{Triple-GNN} = -\frac{1}{|\mathcal{N}|}\sum_{e_i \in \mathcal{N}}\boldsymbol{y}_{e_i}\log{\boldsymbol{p}_{e_i}}
\end{align}
where $\boldsymbol{p}_{e_i} \in \boldsymbol{P}_Q$ denotes the probability of entity $e_i$, the training details for each component are as follows: 
\begin{itemize}
\item For the SGDA, we utilize a training set of approximately 25k entries in the format ($Q$, ($\mathcal{S}$, $\sigma$)). The training configuration includes a learning rate of 1e-4, a batch size of 32, and 2 training epochs, we employing Bfloat16 \cite{kalamkar2019study} precision and the AdamW optimizer \cite{loshchilov2019decoupled}. The maximum input length for the SGDA is set to 2048 tokens. 
\item For the SAGB, the training set consists of about 29k entries in the format ($\mathcal{S}$, $\mathcal{T}^{'}$). It is trained for 2 epochs with a learning rate of 1e-5\footnote{The generation of triples constitutes a format-constrained task, which relies minimally on the prior knowledge of the text-based LLM, so we adopt a smaller learning rate to ensure more stable model training.}, using Bfloat16 precision and the AdamW optimizer. The maximum input length is set to 2048 tokens. 
\item The Triple-GNN is trained for 2 epochs with a learning rate of 1e-5.
\item All training processes were conducted on a single machine with 4 $\times$ NVIDIA H100 GPUs.
\end{itemize}

\textbf{Statistics of Trainable Parameters.} The trainable components in our framework primarily include the SGDA and SAGB models, while each possesses a nominal size of 8B parameters. Additionally, the framework incorporates the Triple-GNN module ($\textbf{\textit{W}}_{\phi_3}\approx$8M parameters) and several auxiliary projection layers ($\textbf{\textit{W}}_{\phi_1}\in \mathbb{R}^{512}$ = 512, $\textbf{\textit{W}}_{\phi_2} \in \mathbb{R}^{3072\times512}\approx$1.5M and $\textbf{\textit{W}}_{\phi_4} \in \mathbb{R}^{1024\times512}\approx$0.5M), the aggregate number of above trainable parameters is about 10M parameters.

\subsection{Structure-to-Query Reverse Generation}
\label{sec:reverse_gen}

The training datasets provide high-quality supervision, however they are limited in scale, which consequently restricts its coverage of the logic and schema diversification encapsulated in the KG. To equip our Semantic-to-Structural Projection pipeline with broader generalization capabilities and cover long-tail relations, we propose a novel Structure-to-Query Reverse Generation strategy, which constructs a large-scale, synthetic instruction-tuning dataset directly from the KG topology.
We elaborate on the procedure in two distinct phases.

\textbf{Phase 1: Reasoning-Path Subgraph Sampling.} For each graph $\mathcal{G}$ in the WebQSP and CWQ datasets, we employ a random walk strategy to obtain a corresponding subgraph $\mathcal{G}_{sub}$ and its associated entity set $\mathcal{N}_{sub}$. Specifically, for subgraph $\mathcal{G}_{sub}$, we randomly mask a subset of nodes to serve as both the target answers and intermediate reasoning entities, replacing them with unified ``[ENTX]'' placeholders consistent with Section \ref{sec:basic-training}, designating the remaining structure, denoted as $\mathcal{G}^{*}_{sub}$, as the evidence subgraph required for reasoning.

\begin{figure*}[t]
    \centering
    \includegraphics[width=\textwidth, trim=2pt 2pt 2pt 2pt, clip]{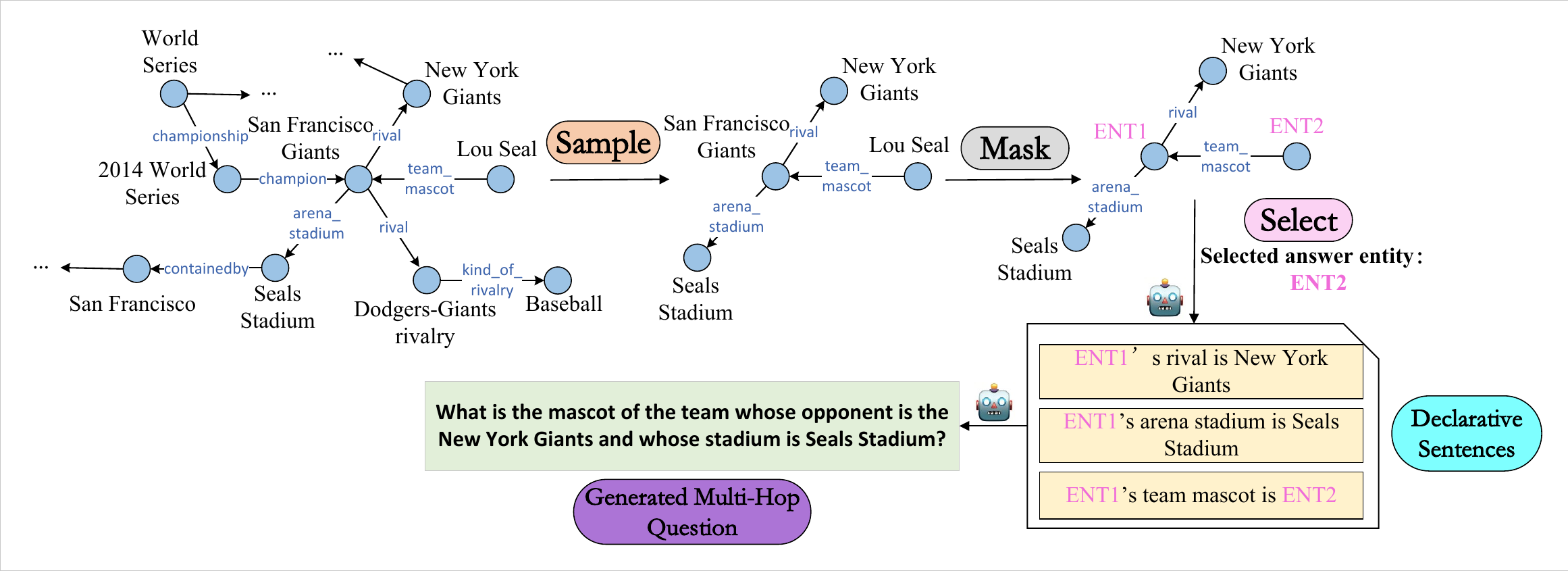}
    \caption{An illustrative example of the Structure-to-Query Reverse Generation pipeline.}
    \label{fig:rgd}
\end{figure*}

\textbf{Phase 2: LLM-Driven Reverse Generation.} We leverage a powerful LLM to generate natural language queries from the sampled subgraphs. Formally, sampled and masked graph $\mathcal{G}^{*}_\text{sub} = \{(e_1, r_1, e_2), (e_2, r_2, \text{``[ENT1]''})\}$ (assuming $e_3$ is masked to serve as the answer entity), we instruct the large language model to generate multi-hop question $Q$ and declarative statements $ \mathcal{S}_\text{sub} $ using prompt \hyperref[fig:p6]{\( \mathcal{P}_6 \)}, $\mathcal{G}_\text{sub}$ and designated answer entity ``[ENT1]'':
\begin{align}
(Q, \mathcal{S}_\text{sub}) \leftarrow \text{LLM}(\mathcal{P}_6, &\mathcal{G}^{*}_\text{sub}, ``\text{[ENT1]}")
\end{align}
where $\mathcal{S}_\text{sub}=\{s_1, s_2, ..., s_n\}$. Considering the complexity of this prompt, which entails a two-step task execution, we enabled the ``thinking mode'' throughout the LLM inference process. We then employ prompt \hyperref[fig:p5]{\( \mathcal{P}_5 \)}, $Q$ and $ \mathcal{S}_\text{sub} $ to generate corresponding answering strategy $\sigma$, following the method described in Appendix \ref{sec:basic-training}. 

Through the LLM-driven Reverse Generation method, we ultimately construct the knowledge graph reverse-generation dataset $\mathcal{D}_\text{syn}$, for every $d_i \in \mathcal{D}_\text{syn}$:
\begin{align}
d_i = (Q^i, \mathcal{A}^i, \mathcal{G}^{i}, \mathcal{N}^{i}_{sub}, \mathcal{G}^{*i}_\text{sub}, \mathcal{S}_\text{sub}^i, \sigma^i)
\end{align}
where $\mathcal{A}^i$ denotes the answer to $Q^i$, corresponding to the masked entity $e_3$ in the example of $ \mathcal{G}^{*}_\text{sub} $. 

For the training of the SGDA, the objective function is defined as follows:
\begin{align}
    \mathcal{L}_{\text{SGDA}}&=  \nonumber\\ - \sum_{i=1}^{|\mathcal{D}|} \sum_{k=1}^{|(\textit{S}^i_{sub}, \sigma^i)|} \log P_{\phi_6}&(y_{i,k} \mid \mathcal{P}_1, Q^i, y_{i,<k})
\end{align}

For the training of the SAGB, the objective function is defined as follows:
\begin{align}
    \mathcal{L}_{\text{SAGB}} &= \nonumber\\ - \sum_{i=1}^{|\mathcal{D}|} \sum_{k=1}^{|\mathcal{G}^{*i}_{sub}|} \log P_{\phi_5}&(y_{i,k} \mid \mathcal{P}_2, \mathcal{S}_{sub}^i, y_{i,<k})
\end{align}

The training objective of the Triple-GNN is consistent with that described in \ref{sec:basic-training}. Regarding labeling, we perform positive and negative sampling within the entity set $\mathcal{N}^{i}_{sub}$, adhering to the following labeling principles:
\begin{align}
\boldsymbol{y}_{e} = \begin{cases} 
\boldsymbol{1}, & e \in \mathcal{N}^i_{sub}, \\
\boldsymbol{0}, & \text{otherwise}.
\end{cases}
\end{align}

For clarity, we present a complete example of the construction process as shown in Figure \ref{fig:rgd}. Ultimately, we constructed dataset $\mathcal{D}_\text{syn}$ comprising about 210k entries. Detailed statistical information regarding $\mathcal{D}_\text{syn}$ is presented in Table \ref{tab:rgd_stat}. This dataset is utilized to train the SGDA, SAGB, and Triple-GNN modules, following the same logic as previously described. We finally incorporate $\mathcal{D}_\text{syn}$ into the the original training set in Appendix \ref{sec:basic-training}. We will validate the impact of the $\mathcal{D}_\text{syn}$ dataset by comparing performance with and without it in the Appendix \ref{sec:additional_exp}. The synthetic dataset $\mathcal{D}_\text{syn}$ will be released alongside the source code.

\subsection{Data ethics} 
We employ LLMs to generate intermediate assertions and multi-hop queries, a potential risk in such synthetic data generation is hallucination. To mitigate this, our Structure-to-Query Reverse Generation method is strictly grounded in sampled subgraphs from the KG. The reasoning paths are pre-determined by the graph structure, ensuring that the generated questions and assertions are logically consistent with the underlying facts. While we have conducted manual quality checks on random samples, we acknowledge that minor semantic noise may persist. Furthermore, given that Knowledge Graphs and LLMs are known to exhibit inherent biases (e.g., geographical, gender, or cultural), our synthetic data—being derived from these sources—may inadvertently propagate such pre-existing biases. Furthermore, the synthetic data constructed via our reverse generation method is derived exclusively from public knowledge graph and standard LLM outputs, these public resources have already been anonymized and sanitized to remove personally identifiable information, thus all training datasets contain no PII or private data.

\begin{table}[t]
\centering
\small
\renewcommand{\arraystretch}{1.15} % 行高
\setlength{\tabcolsep}{2.5pt} 

\begin{tabular}{c|c}
\hline
\textbf{Statistic} & \textbf{Value} \\
\hline
Total Count & 214,733 \\
\hline
Avg Question Length & 49 \\
\hline
Hop=1 & 84,465 \\
\hline
Hop=2 & 65,810 \\
\hline
Hop=3 & 34,210 \\
\hline
Hop$\geq 4$ & 30,248 \\
\hline
\textit{Precision} & 132,391 \\
\hline
\textit{Breadth} & 82,342 \\
\hline
\end{tabular}
\caption{Statistics of the synthetic dataset $\mathcal{D}_\text{syn}$, including total size, hop-count distribution, and strategy distribution.}
\label{tab:rgd_stat}
\end{table}

\section{Additional Experimental Results and Analysis}
\label{sec:additional_exp}

\subsection{The Impact of the Answer Strategy $\sigma$ on Performance}
\label{sec:appendixd1}

\begin{figure*}[t]
    \centering

    \begin{subfigure}{\textwidth}
        \centering
        \includegraphics[width=0.9\textwidth, height=0.19\textheight, trim=2pt 2pt 2pt 2pt, clip]{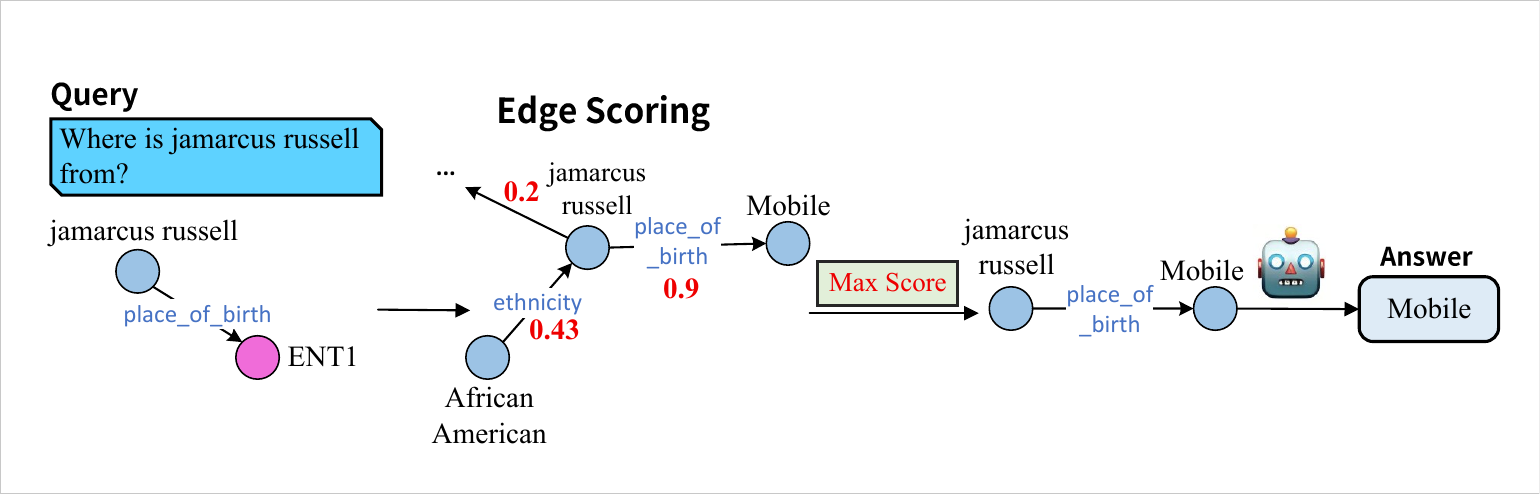}
        \caption{An illustrative example of Greedy Selection, corresponding to the \textit{Precision} answering strategy.}
        \label{fig:pre}
    \end{subfigure}

    \vspace{2mm}

    \begin{subfigure}{\textwidth}
        \centering
        \includegraphics[width=0.9\textwidth, height=0.18\textheight, trim=2pt 2pt 2pt 2pt, clip]{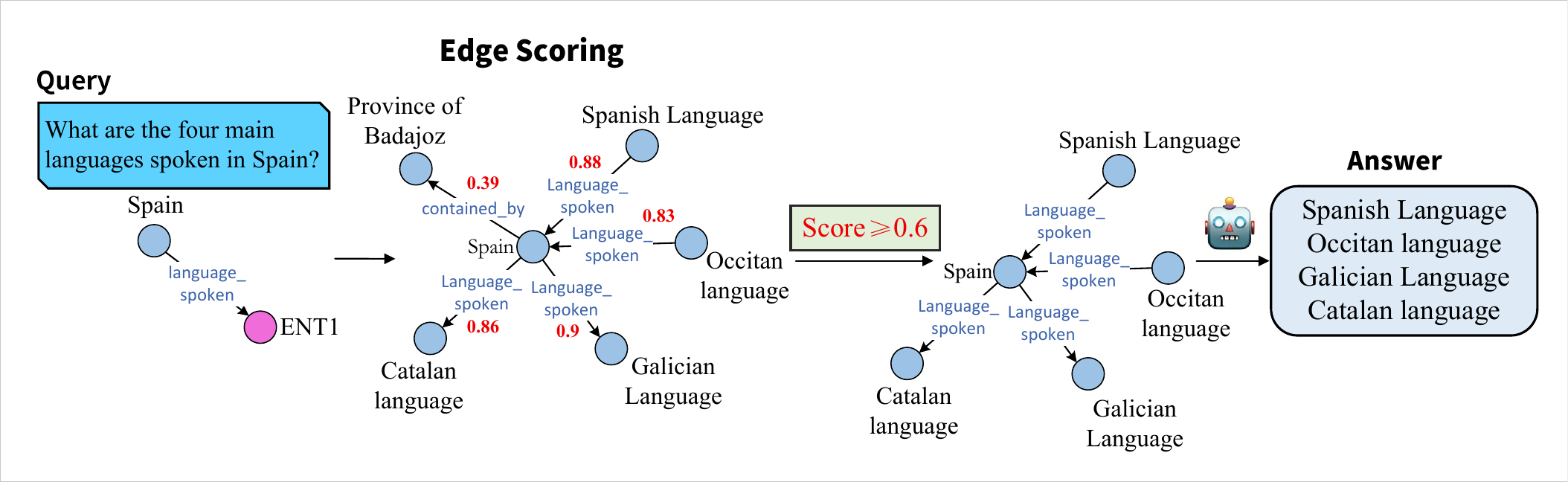}
        \caption{An illustrative example of Threshold-based Selection, corresponding to the \textit{Breadth} answering strategy.}
        \label{fig:brd}
    \end{subfigure}
\end{figure*}

In this section, we investigate the impact of answer strategy differentiation on the experimental results. To illustrate the workflows and distinct behaviors of the search modes under the \textit{Precision} and \textit{Breadth} strategies, we first provide a representative example for Greedy Selection in Figure \ref{fig:pre} and one for Threshold-based Selection in Figure \ref{fig:brd}.

We organize the experiments into three groups: the original Adaptive STEM Strategy, a purely Greedy Selection strategy, and a purely Threshold-based Selection strategy. The experimental results are presented in Table \ref{tab:resp_stra}. 

As shown in the results, the Only Greedy setting exhibits a significant performance decline in the F1 score, this is primarily because greedy search suffers from insufficient evidence recall in multi-answer scenarios, leading to incomplete answers. In contrast, the Only Threshold-based setting does not show a drastic drop and even outperforms the original STEM configuration on both F1 metrics. This is because Threshold-based Selection ensures answer coverage. However, it still underperforms STEM in other metrics. This is attributed to the retrieval of excessive irrelevant evidence, which induces hallucinations. We discuss the efficiency of the two search modes in Appendix \ref{sec:ea}.

\begin{table}[t]
\centering
\footnotesize
\renewcommand{\arraystretch}{1.15}
\setlength{\tabcolsep}{2.5pt} 

\begin{tabular}{l|cc|cc}
\toprule
\hline
\multirow{2}{*}{\textbf{Strategy}} & 
\multicolumn{2}{c|}{\textbf{WebQSP}} & 
\multicolumn{2}{c}{\textbf{CWQ}} \\ \cline{2-5} 

& Hit@1 & $F_1$ Score & Hit@1 & $F_1$ Score \\ 

\hline
Only Greedy & 88.50 & 60.75 & 72.45 & 44.29 \\
Only Threshold-based & 89.36 & 78.54 & 74.53 & 67.18 \\ 
STEM  & 90.94 & 76.18 & 74.09 & 65.33 \\
\hline
\bottomrule
\end{tabular}
\caption{Performance comparison with different response strategies.}
\label{tab:resp_stra}
\end{table}

\subsection{Initial Entity Count $\mathcal{K}$ on Performance}
During the construction of the Global Guidance Subgraph, we perform an initial retrieval of $\mathcal{K}$
entities. In this section, we conduct experiments to evaluate the impact of different $\mathcal{K}$ values, to illustrate this more clearly, we assume $\mathcal{K}$ = $\mid\mathcal{T}\mid*\mathcal{K^{'}}$ and define $\mathcal{K}^{'}$ as the \textbf{Guidance Graph Scale Factor}, we investigate the impact of different values of $\mathcal{K}^{'}$ on performance.

The comparison results of all parameters $\mathcal{K^{'}}$ across different metrics are shown in Figure \ref{fig:k_analysis_hit} (Hit@1) and Figure \ref{fig:k_analysis_f1} (F1). From the results in the tables, it can be seen that values of $\mathcal{K}^{'}$ smaller or larger than 4 both affect the performance. Particularly when $\mathcal{K}^{'}$=1, there is a significant decline in metrics across both datasets. In terms of Hit@1 on both datasets, performance generally declines when $\mathcal{K^{'}}$ is less than 4, and gradually improves as $\mathcal{K^{'}}$ increases, with the CWQ dataset even showing a slightly better result at $\mathcal{K^{'}}$=3 compared to the reported results with $\mathcal{K^{'}}$=4. However, performance drops again when $\mathcal{K^{'}}$ exceeds 4. Regarding F1, a similar declining trend is observed when $\mathcal{K^{'}}$ is less than 4, and simultaneously a moderate decline is observed when $\mathcal{K}^{'}$ exceeds 4, with the exception of the WebQSP dataset. Therefore, $\mathcal{K}^{'}$=4 is selected as our final configuration. Our analysis suggests that a smaller $\mathcal{K^{'}}$ results in a smaller Guidance Graph, which risks missing key entities or question entities, while a larger $\mathcal{K^{'}}$ may introduce low-value entities and mislead the evidence search.  

\pgfplotsset{compat=1.18}

\begin{figure}[t]
    \centering

    \begin{subfigure}[b]{0.45\textwidth}

        \begin{tikzpicture}
            \begin{axis}[
            width=7.5cm, height=6cm, 
            xlabel={Guidance Graph Scale Factor ($\mathcal{K^{'}}$)}, 
            ylabel={Hit@1 (\%)},
            xmin=1, xmax=6,            % X轴范围
            ymin=60, ymax=95,           % Y轴范围 (根据您的数据调整)
            xtick={1, 2, 3, 4, 5, 6},   % X轴显示的刻度点
            ytick={60, 70, 80, 90}, % Y轴显示的刻度点
            grid=major,                 % 显示主要网格
            grid style={dashed,gray!30},% 网格样式：虚线、浅灰
            legend pos=south east,      % 图例位置：右下角 (可选 north west 等)
            nodes near coords,              % 开启数值显示
            nodes near coords style={       % 设置数值样式
                font=\tiny,                 % 字体变小
                yshift=5pt,                 % 向上偏移一点，避免盖住点
                color=black                 % 强制文字颜色为黑
            },
            legend style={              % 图例样式微调
                font=\small,
                draw=none,              % 无边框
                fill=white,             % 背景白
                fill opacity=0.8        % 半透明防遮挡
            },
            cycle list name=exotic,     % 自动循环颜色（可选）
            every axis plot/.append style={thick} % 所有线条加粗
            ]

                \addplot[
                    color=blue,
                    mark=triangle*,
                    mark size=2.5pt
                ]
                coordinates {
                    (1, 88.56)(2, 88.64)(3, 89.59)(4, 90.94)(5, 89.43)(6, 89.66)
                };
                \addlegendentry{WebQSP}
    
                \addplot[
                    color=red,              % 红色突出自己的方法
                    mark=*,                 % 圆点
                    mark size=2.5pt,
                    line width=1.5pt        % 线条更粗一点
                ]
                coordinates {
                    (1, 72.41)(2, 73.05)(3, 74.47)(4, 74.09)(5, 74.05)(6, 73.84)
                };
                \addlegendentry{CWQ}
    
            \end{axis}
        \end{tikzpicture}
        \caption{Impact of Guidance Graph construction scale on performance.}
        \label{fig:k_analysis_hit}
    \end{subfigure}
    
    \begin{subfigure}{0.45\textwidth}
        \centering
        \begin{tikzpicture}
            \begin{axis}[
            width=7.5cm, height=6cm, 
            xlabel={Guidance Graph Scale Factor ($\mathcal{K^{'}}$)}, 
            ylabel={F1 (\%)},
            xmin=1, xmax=6,           
            ymin=50, ymax=80,          
            xtick={1, 2, 3, 4, 5, 6},  
            ytick={50, 60, 70, 80},
            grid=major,                
            grid style={dashed,gray!30},
            legend pos=south east,
            nodes near coords,             
            nodes near coords style={     
                font=\tiny,                
                yshift=5pt,                
                color=black                
            },
            legend style={             
                font=\small,
                draw=none,            
                fill=white,           
                fill opacity=0.8      
            },
            cycle list name=exotic,    
            every axis plot/.append style={thick} 
            ]
    
                \addplot[
                    color=blue,
                    mark=triangle*,
                    mark size=2.5pt
                ]
                coordinates {
                    (1, 73.85)(2, 74.51)(3, 75.06)(4, 76.18)(5, 75.85)(6, 76.42)
                };
                \addlegendentry{WebQSP}
    
                \addplot[
                    color=red,             
                    mark=*,                
                    mark size=2.5pt,
                    line width=1.5pt       
                ]
                coordinates {
                    (1, 61.59)(2, 64.81)(3, 64.40)(4, 65.33)(5, 64.99)(6, 63.65)
                };
                \addlegendentry{CWQ}
    
            \end{axis}
        \end{tikzpicture}
        \caption{Impact of Guidance Graph construction scale on performance.}
        \label{fig:k_analysis_f1}
    \end{subfigure}
\end{figure}

\subsection{Guidance Graph Correction Analysis}
\label{sec:ggca}

While Table \ref{tab:bias_abla} demonstrates that ablating the Guidance Graph significantly degrades overall QA performance, these end-to-end metrics are inevitably confounded by LLM generation stochasticity. To explicitly quantify the Guidance Graph's error-correction capability independent of the LLM, we conduct a pure retrieval evaluation. Specifically, we measure the coverage of ground-truth reasoning paths within the retrieved subgraphs with and without Guidance Graph guidance (using Equation \ref{coverage}). Results are presented in Table \ref{tab:gg_cov}.

\begin{table}[t]
\footnotesize
\centering
\renewcommand{\arraystretch}{1.15}
\setlength{\tabcolsep}{2.5pt} 

\begin{tabular}{l|c|c}
\toprule
\hline
{\textbf{Scoring Bias}} & {\textbf{WebQSP}} & {\textbf{CWQ}} \\

\hline
\textbf{STEM$_{GPT-4o}$}  & 73.68 & 70.39 \\
\textbf{\;\;\;\; w/o $\mathbb{I}_\text{Ent}$ $\&$ $\mathbb{I}_\text{Tri}$}  & 65.07 & 59.8 \\
\textbf{\;\;\;\; w/o $\mathbb{I}_\text{Ent}$}    & 70.25 & 66.68  \\
\textbf{\;\;\;\; w/o $\mathbb{I}_\text{Tri}$} & 68.49 & 65.04  \\ 
\hline
\bottomrule
\end{tabular}
\caption{Coverage rate of ground-truth reasoning paths within retrieved evidence subgraphs. We compare the pure retrieval quality under settings with and without the Guidance Graph. All values are reported as percentages (\%).}
\label{tab:gg_cov}
\end{table}

As shown in Table \ref{tab:gg_cov}, the full Guidance Graph configuration (incorporating both Entity-level and Triple-level biases) achieves the highest retrieval coverage. Quantitatively, the inclusion of Guidance Graph increases coverage from 65.07\% to 73.68\% on WebQSP (an absolute improvement of over 8\%) and from 59.8\% to 70.39\% on CWQ (an increase exceeding 10\%). Conversely, removing both constraints yields the lowest coverage, particularly on the more complex CWQ dataset. This significant drop indicates that without global structural guidance, the subgraph search is highly susceptible to being misled by local semantic features, leading to severe error propagation. Furthermore, ablating either bias individually outperforms the unconstrained variant but remains inferior to the fully constrained variant, confirming their complementary roles.

\subsection{Impact of Global Structural Consistency Bias Values on Performance}
\label{eq:gscb_param}
In this section, we conduct experiments on bias constraints ($\mathbb{I}_\text{Ent}$ and $\mathbb{I}_\text{Tri}$). Specifically, we apply a multiplicative factor of $3/2$ to the scores during initial nodes selection in Equation \ref{entity_bias} and \ref{eq:weighting_func}, and an additive boost of $1/2$ to the scores during the edge search in Equation \ref{eq:t-score} and \ref{eq:triple_bias}. Given that this involves the joint adjustment of two variables, we adopt a grid search strategy. Let $\lambda$ denote the multiplicative factor for the Entity-level bias, and $\tau$ denote the boosting factor for the Triple-level bias. We define the search ranges for $\lambda$ and $\tau$ as follows:
\begin{align}
\lambda &\in \{1.2, 1.5, 1.8, 2.1, 2.4, 2.7 \} \\
\tau &\in \{0.2, 0.5, 0.8, 1.1, 1.4, 1.7, 2.0 \}
\label{eq:param_range}
\end{align} 

Since grid search involves a large number of experimental iterations, we randomly sample 200 examples from WebQSP and CWQ dataset to construct WebQSP$_\text{sub}$ and CWQ$_\text{sub}$, and conduct experiments on them. We fix one variable and search for the optimal setting of the other, all experimental results are presented in Figure \ref{fig:lambda_search} and Figure \ref{fig:tau_search}.

From the experimental results, we can conclude that when both $\lambda$ and $\tau$ are relatively small, performance drops significantly. In particular, with $\lambda$ = 1.2, the WebQSP score decreases by about 3\%, as $\lambda$ increases from 1.5 onward, scores generally improve and remain stable thereafter. A similar trend is observed for $\tau$: with $\tau$ = 0.2, the performance also deteriorates. Starting from $\tau$ = 0.5, the scores improve and stay stable. Since larger parameters do not bring significant further improvements, we select the relatively low values $\lambda$ = 1.5 and $\tau$ = 0.5 as our final configuration.

\textbf{Our Analysis} The results suggests that excessively high parameter values lead to an over-reliance on the Guidance Graph. We posit that during query-based Guidance Graph building, the GNN may inadvertently incorporate edges with low relevance. These edges, while structurally connected, often contribute little to the actual reasoning process—a phenomenon we term ``\textbf{Structural Over-Interpretation}''. This limitation elucidates why the Guidance Graph cannot serve as the final reasoning subgraph in isolation and necessitates a subsequent refinement via semantic search. Fundamentally, STEM represents a synergy between logical reasoning (structure) and semantic matching (content). It achieves an optimal equilibrium, avoiding over-reliance on either modality while leveraging the indispensable strengths of both.

\begin{figure}[t]
    \centering

    \begin{subfigure}{.48\textwidth}
        \centering
        \begin{tikzpicture}
            \begin{axis}[
            width=7.5cm, height=6cm, 
            enlarge x limits=0.06,
            xlabel={Multiplicative factor $\lambda$}, 
            ylabel={F1 (\%)},
            xmin=1.2, xmax=2.7,           
            ymin=40, ymax=80,          
            xtick={1.2, 1.5, 1.8, 2.1, 2.4, 2.7},  
            ytick={40, 50, 60, 70, 80},
            grid=major,                
            grid style={dashed,gray!30},
            legend pos=south east,
            nodes near coords,             
            nodes near coords style={     
                font=\tiny,                
                yshift=5pt,                
                color=black                
            },
            legend style={             
                font=\small,
                draw=none,            
                fill=white,           
                fill opacity=0.8      
            },
            cycle list name=exotic,    
            every axis plot/.append style={thick} 
            ]
    
                \addplot[
                    color=gray,
                    dashed,
                    mark=square*,
                    mark size=2.5pt
                ]
                coordinates {
                    (1.2, 67.15)(1.5, 70.18)(1.8, 70.30)(2.1, 70.54)(2.4, 70.12)(2.7, 70.35)
                };
                \addlegendentry{WebQSP (sub)}
    
                \addplot[
                    color=orange,  
                    dashed,
                    mark=diamond*,                
                    mark size=2.5pt,
                    line width=1.5pt       
                ]
                coordinates {
                    (1.2, 52.71)(1.5, 54.22)(1.8, 54.16)(2.1, 53.19)(2.4, 54.10)(2.7, 53.98)
                };
                \addlegendentry{CWQ (sub)}
    
            \end{axis}
        \end{tikzpicture}
        \caption{Performance comparison with different $\lambda$. Due to the constraints of the controlled variable method, the value of $\tau$ is set to 0.2 for all experiments.}
        \label{fig:lambda_search}
    \end{subfigure}
    \begin{subfigure}{0.45\textwidth}
        \centering
        \begin{tikzpicture}
            \begin{axis}[
            width=7.5cm, height=6cm, 
            enlarge x limits=0.06,
            xlabel={Boosting factor $\tau$}, 
            ylabel={F1 (\%)},
            xmin=0.2, xmax=2.0,           
            ymin=40, ymax=80,          
            xtick={0.2, 0.5, 0.8, 1.1, 1.4, 1.7, 2.0},  
            ytick={40, 50, 60, 70, 80},
            grid=major,                
            grid style={dashed,gray!30},
            legend pos=south east,
            nodes near coords,             
            nodes near coords style={     
                font=\tiny,                
                yshift=5pt,                
                color=black                
            },
            legend style={             
                font=\small,
                draw=none,            
                fill=white,           
                fill opacity=0.8      
            },
            cycle list name=exotic,    
            every axis plot/.append style={thick} 
            ]
    
                \addplot[
                    color=blue,
                    dashed,
                    mark=square*,
                    mark size=2.5pt
                ]
                coordinates {
                    (0.2, 67.15)(0.5, 68.90)(0.8, 69.49)(1.1, 68.37)(1.4, 68.45)(1.7, 69.02)(2.0, 69.29)
                };
                \addlegendentry{WebQSP (sub)}
    
                \addplot[
                    color=green,  
                    dashed,
                    mark=diamond*,                
                    mark size=2.5pt,
                    line width=1.5pt       
                ]
                coordinates {
                    (0.2, 52.71)(0.5, 55.71)(0.8, 55.94)(1.1, 55.68)(1.4, 54.76)(1.7, 54.93)(2.0, 55.16)
                };
                \addlegendentry{CWQ (sub)}
    
            \end{axis}
        \end{tikzpicture}
        \caption{Performance comparison with different $\tau$. Due to the constraints of the controlled variable method, the value of $\lambda$ is set to 1.2 for all experiments.}
        \label{fig:tau_search}
    \end{subfigure}
\end{figure}

\subsection{Impact of Reverse Generation Data}
We evaluate the impact of Structure-to-Query training set reverse generation on model performance. We first designed two comparative settings: a standard dataset, denoted as $\mathcal{D}_\text{std}$, constructed solely from the training splits of WebQSP and CWQ; and an augmented dataset, denoted as $\mathcal{D}_\text{aug}$, which combines the $\mathcal{D}_\text{std}$ with the synthetic data $\mathcal{D}_\text{syn}$ produced in \ref{sec:reverse_gen}. We trained two separate sets of SGDA, SAGB, and Triple-GNN modules using these respective datasets and conducted comparative experiments. Two specific experiments were designed: \textbf{1. Schema Generation Quality:} We employed the SGDA and SAGB modules to construct schema graphs for queries in the WebQSP and CWQ test sets. The quality of these graphs was then evaluated by measuring precision and recall against the ground-truth reasoning paths. \textbf{2. End-to-End QA Performance:} We integrated SGDA, SAGB, and Triple-GNN into the complete STEM retrieval framework and evaluated the overall question-answering performance on both test sets.

\begin{table}[t]
\centering
\footnotesize
\renewcommand{\arraystretch}{1.15}
\setlength{\tabcolsep}{2.5pt} 

\begin{tabular}{l|cc|cc}
\toprule
\hline
\multirow{2}{*}{\textbf{Dataset}} & 
\multicolumn{2}{c|}{\textbf{WebQSP}} & 
\multicolumn{2}{c}{\textbf{CWQ}} \\ \cline{2-5} 

& Hit@1  & $F_1$ Score  & Hit@1 & $F_1$ Score \\ 

\hline
$\mathcal{D}_{\text{std}}$ & 86.35 & 74.17 & 67.03 & 58.78 \\
$\mathcal{D}_{\text{aug}}$ & 90.94 & 76.18 & 74.09 & 65.33 \\
\hline
\bottomrule
\end{tabular}
\caption{Ablation study on reverse generation data: comparison of multi-hop QA results.}
\label{tab:e2e_QA}
\end{table}

\subsubsection{Graph Evaluation}
To evaluate schema graph generation quality, we first define the Precision and Recall metrics for the generated schema graphs. Given a question $Q$, let $\mathcal{R}$ denote the ground-truth reasoning path provided in the test set, and $\mathcal{G}_{sch}$ denote the schema graph generated by our trained Semantic-to-Structural Projection pipeline. We calculate Precision, Recall, and F1 score as follows:
\begin{gather}
\text{Precision} = \frac{|\mathcal{R} \cap \mathcal{G}_{sch}|}{|\mathcal{G}_{sch}|}, \\
\text{Recall} = \frac{|\mathcal{R} \cap \mathcal{G}_{sch}|}{|\mathcal{R}|}, \\
\text{F1} = \frac{2 \cdot \text{Precision} \cdot \text{Recall}}{\text{Precision} + \text{Recall}}
\end{gather}

The experimental results are presented in Figure \ref{fig:rg_prf}. It is evident that incorporating the $\mathcal{D}_\text{aug}$ data leads to significant improvements in schema generation Precision, Recall, and F1 scores across both test sets. Notably, on WebQSP, the inclusion of $\mathcal{D}_\text{aug}$ yields a Recall increase of approximately 15\% and an F1 improvement exceeding 14\%. Similarly, the CWQ dataset witnesses a marked 15\% rise in Precision and a 12\% gain in Recall. Collectively, these results demonstrate that the incorporation of $\mathcal{D}_\text{aug}$ significantly bolsters the model's capability to perform logical perception for complex queries.

\begin{figure}[t]
    \centering
    \begin{tikzpicture}
        \begin{axis}[
            ybar,
            bar width=7pt,
            width=\columnwidth,
            height=7.2cm,
            ylabel={Precision / Recall / F1 (\%)},
            % --- 内部代号极简纯字母化 ---
            symbolic x coords={QSP1, QSP2, CWQ1, CWQ2},
            xtick=data,
            % --- 使用 \mathrm 代替 \text ---
            xticklabels={
                {WebQSP / $\mathcal{D}_{\mathrm{std}}$},
                {WebQSP / $\mathcal{D}_{\mathrm{aug}}$},
                {CWQ / $\mathcal{D}_{\mathrm{std}}$},
                {CWQ / $\mathcal{D}_{\mathrm{aug}}$}
            },
            ymin=20, ymax=100,
            enlarge x limits=0.25,
            xticklabel style={
                font=\small,
                rotate=35,
                anchor=north east,
                inner sep=1pt
            },
            legend style={
                at={(0.7, 1.15)},
                anchor=north,
                legend columns=-1,
                draw=none,
                font=\small
            },
            grid=major,
            grid style={dashed,gray!30},
            nodes near coords,
            nodes near coords style={
                font=\tiny,
                rotate=90,
                anchor=west
            }
        ]
            % --- 数据点 ---
            \addplot[fill=cyan!40, draw=black] coordinates {
                (QSP1, 76.62) (QSP2, 89.25) (CWQ1, 43.55) (CWQ2, 58.78)
            };
            \addplot[fill=orange!50, draw=black] coordinates {
                (QSP1, 71.93) (QSP2, 87.49) (CWQ1, 45.32) (CWQ2, 57.86)
            };
            \addplot[fill=red!50, draw=black] coordinates {
                (QSP1, 74.20) (QSP2, 88.36) (CWQ1, 44.41) (CWQ2, 58.31)
            };
            \legend{Precision, Recall, F1}
            
        \end{axis}
    \end{tikzpicture}
    \caption{Ablation study on reverse generation data: comparison of schema graph P/R/F1 metrics.}
    \label{fig:rg_prf}
\end{figure}

\subsubsection{End-to-End QA Performance}
We constructed complete STEM retrieval-QA pipelines using modules trained on the two respective datasets and conducted a comparative evaluation, with results shown in Table \ref{tab:e2e_QA}. The end-to-end tests reveal that the pipeline trained with $\mathcal{D}_\text{aug}$ dataset consistently achieves higher overall scores than the one trained with $\mathcal{D}_\text{aug}$ dataset. Notably, the improvement margin on CWQ is more significant, with Hit@1 increasing by over 7\%. This indicates that the performance benefits yielded by the Structure-to-Query Reverse Generation method are amplified in complex reasoning scenarios involving a higher number of hops, underscoring the critical importance of query planning in multi-hop reasoning tasks.

\begin{table}[t]
\centering
\footnotesize
\renewcommand{\arraystretch}{1.15} % 行高
\setlength{\tabcolsep}{2.5pt} 

\begin{tabular}{c|cccc}
\hline
{\textbf{Dataset}} & {\textbf{Ans} = 1} & {\textbf{Ans} $\in$ [2,4]} & {\textbf{Ans} $\in$ [5,9]} & {\textbf{Ans} $\geq$ 10} \\

\hline
\textbf{WebQSP}    &  3.1 & 4.03 & 5.59 & 8.81 \\
\textbf{CWQ}  & 3.41 & 3.68 & 6.13 & 9.03 \\
\hline
\end{tabular}
\caption{Detailed average reasoning time (s) partitioned by answer count intervals.}
\label{tab:per_speed_ans_count}
\end{table}

\begin{table}[t]
\centering
\footnotesize
\renewcommand{\arraystretch}{1.15} % 行高
\setlength{\tabcolsep}{2.5pt} 

\begin{tabular}{c|cc}
\hline
{\textbf{Dataset}} & {\textbf{Ans} = 1} & {\textbf{Ans} $\geq$ 2} \\

\hline
\textbf{WebQSP}    &  96.5 & 93.31 \\
\textbf{CWQ}  & 93.91 & 91.62 \\
\hline
\end{tabular}
\caption{Strategy generation accuracy (\%) of SGDA across different test set splits. We categorize questions into two groups based on answer cardinality: those with a single answer (count $= 1$) and those with answer counts $\geq 2$. For the former, accuracy is measured by the proportion of generated \textit{Precision} strategies, while for the latter, it is measured by the proportion of \textit{Breadth} strategies. }
\label{tab:sgda_acc}
\end{table}

\begin{table}[t]
\centering
\renewcommand{\arraystretch}{1.15} % 保持行高
\setlength{\tabcolsep}{12pt} % 稍微增加列间距，让单栏表格看起来不那么挤

\begin{tabular}{c|cc} % 注意：原代码是 c|cccc 但实际只有3列，这里修正为 c|cc
\hline
\textbf{Dataset} & \textit{Precision} & \textit{Breadth} \\
\hline
\textbf{WebQSP} & 4.91 & 8.42 \\
\textbf{CWQ}    & 4.35 & 7.98 \\
\hline
\end{tabular}
\caption{Latency comparison (s) of different answering strategies.}
\label{tab:pre_brd}
\end{table}

\begin{figure}[t]
    \centering
    \begin{tikzpicture}
        \begin{axis}[
            width=7.6cm, height=6cm,
            enlarge x limits=0.06,
            xlabel={Beam size $\mathcal{B}$},
            ylabel={F1 (\%)},
            xmin=1, xmax=7,
            ymin=30, ymax=85,
            xtick={1, 2, 3, 4, 5, 6, 7},
            ytick={30, 40, 50, 60, 70, 80},
            grid=major,
            grid style={dashed,gray!30},
            legend pos=south east,
            nodes near coords,
            nodes near coords style={
                font=\tiny,
                yshift=5pt,
                color=black
            },
            legend style={
                font=\small,
                draw=none,
                fill=white,
                fill opacity=0.8
            },
            cycle list name=exotic,
            every axis plot/.append style={thick}
        ]
            % --- WebQSP 数据线 ---
            \addplot[
                color=gray,
                dashed,
                mark=square*,
                mark size=2.5pt
            ]
            coordinates {
                (1, 64.27)(2, 67.78)(3, 69.84)(4, 76.18)(5, 76.82)(6, 76.45)(7, 77.09)
            };
            \addlegendentry{WebQSP}

            % --- CWQ 数据线 ---
            \addplot[
                color=orange,
                dashed,
                mark=diamond*,
                mark size=2.5pt,
                line width=1.5pt
            ]
            coordinates {
                (1, 45.28)(2, 46.59)(3, 53.43)(4, 65.69)(5, 64.56)(6, 66.02)(7, 67.57)
            };
            \addlegendentry{CWQ}
        \end{axis}
    \end{tikzpicture}
    \caption{Performance comparison with different beam size $\mathcal{B}$.}
    \label{fig:beam_size}
\end{figure}

\subsection{Impact of SGDA Beam Size \( \mathcal{B} \) on Performance}

Given the inherent structural complexity of KG schemas, the SGDA employs beam search to generate multiple candidate sequences of atomic relational assertions simultaneously. In this section, we investigate the impact of the beam size \( \mathcal{B} \) on model performance. The experimental results are illustrated in Figure \ref{fig:beam_size}.

It is evident that with a small beam size, retrieval failures often arise from the insufficiency of generated assertions. Specifically, at \( \mathcal{B} =1 \), the model yields only the single most probable assertion; however, since queries of the same semantic type do not invariably map to a single identical pattern, retrieving a diverse set of multi-pattern assertions significantly enhances the structural hit rate of the evidence subgraph. Consequently, performance improves significantly as \( \mathcal{B} =1 \) gradually increases. This trend is particularly pronounced on CWQ. Conversely, increasing \( \mathcal{B} \) beyond 4 yields only marginal performance gains. Consequently, to balance retrieval accuracy with the computational complexity induced by processing excessive assertions, we adopt \( \mathcal{B} =4\) for this work.

\subsection{Typical Case Study}

In this section, we present a qualitative analysis of STEM's capabilities in query understanding, logical planning, schema graph construction, and retrieval through representative case studies. We selected diverse examples from both the WebQSP and CWQ datasets to illustrate various performance characteristics.

We begin with the WebQSP dataset, examining cases C1 (Table \ref{tab:c1}), C2 (Table \ref{tab:c2}), and C3 (Table \ref{tab:c3}). Case C1 (Table \ref{tab:c1}) exemplifies the ``schema hallucination'' challenge discussed in Section \ref{sec:intro}. As shown in the table, the SGDA module successfully mapped the query semantic ``airport to fly into'' to the concept of ``nearby airport'', guiding the SAGB to construct a schema graph consistent with KG facts. We also observe that all four sets of assertions were mapped to an identical schema graph, demonstrating the SAGB's structural consistency in handling such queries. Furthermore, the SGDA correctly predicted the \textit{Breadth} strategy, facilitating successful retrieval of all subgraphs and comprehensive recall of the two answers.

Regarding Case C2 (Table \ref{tab:c2}), based on diverse assertions, the SAGB constructed multiple candidate schema graphs, retrieving answer-irrelevant yet structurally similar evidence subgraphs (e.g., (``Beech Street Historic District'', ``location.location.containedby'', ``Texarkana, Arkansas'')). However, leveraging the LLM's precise discriminative capability, the system successfully identified and selected the correct evidence graph: (``Texarkana, Arkansas'', ``location.hud\_county\_place.county'', ``Miller County'').

Finally, in Case C3 (Table \ref{tab:c3}), the SGDA successfully bridged the semantic gap by transforming the phrase ``style of music'' into the assertion term ``music genre''. This alignment enabled the accurate construction of the schema graph and the subsequent retrieval of supporting evidence.

Turning to the CWQ dataset, we analyze cases C4 (Table \ref{tab:c4}), C5 (Table \ref{tab:c5}), C6 (Table \ref{tab:c6}) and C7 (Table \ref{tab:c7}). The increased hop-count complexity inherent in CWQ inevitably exacerbates the potential for reasoning errors. In Case C4 (Table \ref{tab:c4}), although structurally similar schema graphs were constructed from different assertions, the relation names exhibited significant diversity, and the SAGB incorrectly mapped the relation to symbol ``sports.school\_sports\_team.team''. As indicated in the ``Retrieved'' row, the ground-truth relation symbol is ``sports.school\_sports\_team.school''. However, due to the high semantic similarity retained between the predicted and actual relations, the system still achieved accurate evidence retrieval. Simultaneously, the reasoning model successfully derived the final answer from the retrieved evidence subgraph.

Case C5 (Table \ref{tab:c5}) demonstrates the SGDA's robust divergent reasoning capability when decomposing multi-hop queries. While assertions (1) represents a generic logical form, the structural complexity of assertions (3) stems from the SGDA's comprehensive grasp of KG schemas, exemplifying high-quality factual alignment. Consequently, the schema graph derived from assertions (3) accurately retrieved the target evidence subgraph. In contrast, the retrieval results for assertions (1) and (2) were rejected by the reasoning model due to the absence of critical information regarding the movie ``Forrest Gump'' (the question entity given in the test set) thereby ensuring the output of the correct evidence.

Case C6 (Table \ref{tab:c6}) demonstrates a failure case of the SGDA. The assertions (1) (``Corfu's official language is [ENT1]'') resulted in a retrieval failure due to the absence of corresponding relation within the KG. In contrast, assertions (3) (``Corfu is an administrative division of [ENT1]'', ``[ENT1]’s official language is [ENT2]'') successfully aligned accurately with the underlying facts, ultimately retrieving all correct answers. Meanwhile, assertions (2), utilizing the relation ``location.location.containedby'', retrieved an irrelevant subgraph. 

Finally, Case C7 (Table \ref{tab:c7}) presents an instance of assertion conversion error, where the relationship between ``Brussels'' and the ``European Union'' was misidentified as ``location.location.containedby'' instead of the correct ``organization.organization.founders''. Nevertheless, the Triple-GNN successfully identified the ``European Union'' as a critical node during the construction of the Guidance Graph. Consequently, the correct triple (``European Union'', ``organization.organization.founders'', ``Belgium'') was included in the Guidance Graph, which applied a positive bias to this edge during the search phase. This structural prior effectively corrected the semantic deviation, ensuring the successful recall of the evidence subgraph\footnote{The correct triple (``European Union'', ``organization.organization.founders'', ``Belgium'') would not have been selected as the top-ranked edge under a pure semantic matching regime due to the significant semantic divergence between the relations ``founders'' and ``containedby'', which yielded a low similarity score of 0.64. However, by incorporating the consistency bias, the score of this valid edge was successfully boosted to 1.14 and become the final selected edge.}. Although the relation ``organization.membership\_organization.members'' triggered the retrieval of an irrelevant subgraph, the presence of the entity ``Brussels'' successfully prevented the model from being misled.

Collectively, these findings demonstrate STEM's robust resilience against schema inconsistency, manifested in three key aspects:
\begin{itemize}
\item \textbf{Multiple Planning Hypotheses}: Faced with diverse KG structures, the SGDA generates multiple candidate decomposition plans, significantly increasing the hit rate for the correct knowledge structure.
\item \textbf{Fuzzy Semantic Matching}: In cases of relation symbol mismatch caused by SAGB, the search mechanism compensates via semantic similarity, ensuring successful evidence recall provided that the semantics remain proximate.
\item \textbf{Global Structural Guidance}: For semantically distant relations between schema graph and true logic in KG, the Guidance Graph-derived consistency bias incorporates global structural priors to prioritize potentially optimal edges, thereby safeguarding the correctness of each search step.
\end{itemize}

\begin{table*}[htbp]
\centering

\footnotesize
\renewcommand{\arraystretch}{1.4}
\begin{tabular}{p{2.5cm}|p{11.5cm}}
\hline

\textbf{Question} & 
which airport to fly into rome \\
\hline

\textbf{Assertions} & 
\begin{minipage}[t]{\linewidth}
1. ("rome's nearby airport is [ENT1]",) \\
2. ("the airport near rome is [ENT1].",) \\
3. ("rome is served by a nearby airport, [ENT1].",) \\
4. ("[ENT1] is a nearby airport for rome.",)
\end{minipage} \\
\hline

\textbf{Strategy} & 
\begin{minipage}[t]{\linewidth}
Breadth
\end{minipage} \\
\hline

\textbf{Schema Graphs} & 
\begin{minipage}[t]{\linewidth}
1. [("rome", "location.location.nearby\_airports", "[ENT1]")]
\end{minipage} \\

\hline

\textbf{Retrieved} & 
\begin{minipage}[t]{\linewidth}
1. [("Rome", "location.location.nearby\_airports", "Ciampino–G. B. Pastine International \text{\;\;\;\;}Airport")]\\
2. [("Rome", "location.location.nearby\_airports", "Leonardo da Vinci–Fiumicino Airport")]
\end{minipage} \\

\hline
\raggedright
\textbf{Ground Truth (2 items)} & 
Ciampino–G. B. Pastine International Airport, Leonardo da Vinci–Fiumicino Airport \\

\hline

\textbf{Output Answer} & 
Ciampino - G. B. Pastine International Airport and Leonardo da Vinci – Fiumicino Airport. \\
\hline
\end{tabular}
\caption{Case study C1: Interpretability analysis on the WebQSP dataset.}
\label{tab:c1}
\end{table*}

\begin{table*}[htbp]
\centering

\footnotesize
\renewcommand{\arraystretch}{1.4}
\begin{tabular}{p{2.5cm}|p{11.5cm}}
\hline

\textbf{Question} & 
what county is texarkana arkansas in \\
\hline

\textbf{Assertions} & 
\begin{minipage}[t]{\linewidth}
1. ("texarkana arkansas is a country of [ENT1]",) \\
2. ("texarkana, arkansas is a country within [ENT1].",) \\
3. ("texarkana arkansas is part of the country [ENT1].",) \\
4. ("the country to which texarkana arkansas belongs is [ENT1].",)
\end{minipage} \\
\hline

\textbf{Strategy} & 
\begin{minipage}[t]{\linewidth}
Precision
\end{minipage} \\
\hline

\textbf{Schema Graphs} & 
\begin{minipage}[t]{\linewidth}
1. [("texarkana arkansas", "location.location.containedby", "[ENT1]")]\\
2. [("texarkana arkansas", "location.hud\_county\_place.county", "[ENT1]")]\\
3. [("texarkana arkansas", "location.administrative\_division", "[ENT1]")]\\
\end{minipage} \\

\hline

\textbf{Retrieved} & 
\begin{minipage}[t]{\linewidth}
\raggedright
1. [("Beech Street Historic District", "location.location.containedby", "Texarkana, \text{\;\;\;\;}Arkansas")]\\
2. [("texarkana, arkansas", "location.hud\_county\_place.county", "Miller County")]\\
3. [("Arkansas","location.administrative\_division.country","United States of America")]
\end{minipage} \\

\hline

\textbf{Ground Truth} & 
Miller County \\

\hline

\textbf{Output Answer} & 
Miller County \\
\hline
\end{tabular}
\caption{Case study C2: Interpretability analysis on the WebQSP dataset.}
\label{tab:c2}
\end{table*}

\begin{table*}[htbp]
\centering

\footnotesize
\renewcommand{\arraystretch}{1.4}
\begin{tabular}{p{2.5cm}|p{11.5cm}}
\hline

\textbf{Question} & 
what style of music did bessie smith perform \\
\hline

\textbf{Assertions} & 
\begin{minipage}[t]{\linewidth}
1. ("bessie smith's music genre is [ENT1]",) \\
2. ("the music genre of bessie smith is [ENT1].",) \\
3. ("bessie smith’s genre of music is [ENT1].",) \\
4. ("[ENT1] is the music genre associated with bessie smith.",)
\end{minipage} \\
\hline

\textbf{Strategy} & 
\begin{minipage}[t]{\linewidth}
Precision
\end{minipage} \\
\hline

\textbf{Schema Graphs} & 
\begin{minipage}[t]{\linewidth}
1. [("bessie smith", "music.artist.genre", "[ENT1]")]
\end{minipage} \\

\hline

\textbf{Retrieved} & 
\begin{minipage}[t]{\linewidth}
1. [("Bessie Smith", "music.artist.genre", "Jazz")]
\end{minipage} \\

\hline

\textbf{Ground Truth} & 
Jazz \\

\hline

\textbf{Output Answer} & 
Jazz \\
\hline
\end{tabular}
\caption{Case study C3: Interpretability analysis on the WebQSP dataset.}
\label{tab:c3}
\end{table*}

\begin{table*}[htbp]
\centering

\footnotesize
\renewcommand{\arraystretch}{1.4}
\begin{tabular}{p{2.5cm}|p{11.5cm}}
\hline

\textbf{Question} & 
What educational institution with men's  sports team named Wisconsin Badgers did Russell Wilson go to? \\
\hline

\textbf{Assertions} & 
\begin{minipage}[t]{\linewidth}
1. ("Wisconsin Badgers is a school sports team of [ENT1].", "Russell Wilson's educational \text{\;\;\;\;}institution is [ENT1].") \\
2. ("The school sports team known as the Wisconsin Badgers belongs to [ENT1].", "The \text{\;\;\;\;}educational institution that Russell Wilson attended is [ENT1].") \\
3. ("[ENT1]’s official school sports team is called the Wisconsin Badgers.", "Russell Wilson's \text{\;\;\;\;}educational institution is [ENT1].") \\
4. ("[ENT1] is the institution that fields the Wisconsin Badgers sports team.", "Russell \text{\;\;\;\;}Wilson received his education at [ENT1].")
\end{minipage} \\
\hline

\textbf{Strategy} & 
\begin{minipage}[t]{\linewidth}
Precision
\end{minipage} \\
\hline

\textbf{Schema Graphs} & 
\begin{minipage}[t]{\linewidth}
1.[("Wisconsin Badgers", "sports.sports\_league.teams", "[ENT1]"), ("Russell Wilson", "edu-\text{\;\;\;\;}cation.education.institution", "[ENT1]")] \\
2.[("Wisconsin Badgers", "sports.school\_sports\_team.team", "[ENT1]"), ("Russell Wilson", \text{\;\;\;\;}"education.education.institution", "[ENT1]")] \\
3.[("Wisconsin Badgers", "sports.sports\_league.teams", "[ENT1]"), ("[ENT1]", "edu-\text{\;\;\;\;}cation.education.student", "Russell Wilson")]
\end{minipage} \\

\hline

\textbf{Retrieved} & 
\begin{minipage}[t]{\linewidth}
\raggedright
1.[("Wisconsin Badgers men's basketball", "sports.school\_sports\_team.school", "University \text{\;\;\;\;}of Wisconsin-Madison"), ("Russell Wilson", "education.education.institution", \text{\;\;\;\;}"University of Wisconsin-Madison")] \\
2.[("Wisconsin Badgers", "education.athletics\_brand.teams", "Wisconsin Badgers women's \text{\;\;\;\;}ice hockey")
, ("University of Wisconsin-Madison", \text{\;\;\;\;}"education.educational\_institution.sports\_teams", "Wisconsin Badgers women's ice \text{\;\;\;\;}hockey")] \\
3.[("m.0hpny0z", "education.education.student", "Russell Wilson"),
\text{\;\;\;\;}("m.0hpny0z","education.education.degree","Bachelor of Arts")]
\end{minipage} \\

\hline

\textbf{Ground Truth} & 
University of Wisconsin-Madison \\

\hline

\textbf{Output Answer} & 
University of Wisconsin-Madison \\
\hline
\end{tabular}
\caption{Case study C4: Interpretability analysis on the CWQ dataset.}
\label{tab:c4}
\end{table*}

\begin{table*}[htbp]
\centering

\footnotesize
\renewcommand{\arraystretch}{1.4}
\begin{tabular}{p{2.5cm}|p{11.5cm}}
\hline

\textbf{Question} & 
What actor played the a kid in the movie with a character named Jenny's Father? \\
\hline

\textbf{Assertions} & 
\begin{minipage}[t]{\linewidth}
\raggedright
1. ("Jenny’s father is a movie character in [ENT1].", "[ENT2] performs a role in the \text{\;\;\;\;}production [ENT1].") \\
2. ("Jenny’s father is a character in [ENT1].", "[ENT2] appears as an actor in [ENT1].") \\
3. ("Jenny’s father is a character in movie [ENT1].", "[ENT2] is a character in [ENT1].", \text{\;\;\;\;}"[ENT3] portrayed [ENT2] in the film.")
\end{minipage} \\
\hline

\textbf{Strategy} & 
\begin{minipage}[t]{\linewidth}
Precision
\end{minipage} \\
\hline

\textbf{Schema Graphs} & 
\begin{minipage}[t]{\linewidth}
\raggedright
1.[("Jenny's Father", "film.performance.character", "[ENT1]"), ("[ENT2]", \text{\;\;\;\;}"film.performance.actor", "[ENT1]")] \\
2.[("Jenny's Father", "film.film\_character.portrayed\_in\_films", "[ENT1]"), ("[ENT2]", \text{\;\;\;\;}"film.film\_character.portrayed\_in\_films", "[ENT1]"), ("[ENT2]", \text{\;\;\;\;}"film.performance.actor", "[ENT3]")]
\end{minipage} \\

\hline

\textbf{Retrieved} & 
\begin{minipage}[t]{\linewidth}
\raggedright
1.[("m.0y54dnx", "film.performance.character", "Jenny's Father"), \text{\;\;\;\;}("m.0y54dnx","film.performance.actor","Kevin Mangan")] \\
2.[("Jenny's Father", "film.film\_character.portrayed\_in\_films", "Forrest Gump"), ("Forrest \text{\;\;\;\;}Gump", "film.film\_character.portrayed\_in\_films", "m.02xgww5"), ("m.02xgww5", \text{\;\;\;\;}"film.performance.actor", "Michael Connor Humphreys")]
\end{minipage} \\

\hline

\textbf{Ground Truth} & 
Michael Connor Humphreys \\

\hline

\textbf{Output Answer} & 
Michael Connor Humphreys \\
\hline
\end{tabular}
\caption{Case study C5: Interpretability analysis on the CWQ dataset.}
\label{tab:c5}
\end{table*}

\begin{table*}[htbp]
\centering

\footnotesize
\renewcommand{\arraystretch}{1.4}
\begin{tabular}{p{2.5cm}|p{11.5cm}}
\hline

\textbf{Question} & 
People from the country that contains Corfu speak what language? \\
\hline

\textbf{Assertions} & 
\begin{minipage}[t]{\linewidth}
1. ("Corfu's official language is [ENT1].",) \\
2. ("Corfu is belong to [ENT1].", "[ENT1]'s official language is [ENT2].") \\
3. ("Corfu is an administrative division of [ENT1].", "[ENT1]'s official language is [ENT2].")
\end{minipage} \\
\hline

\textbf{Strategy} & 
\begin{minipage}[t]{\linewidth}
Breadth
\end{minipage} \\
\hline

\textbf{Schema Graphs} & 
\begin{minipage}[t]{\linewidth}
\raggedright
1.[("Corfu", "location.country.official\_language", "[ENT1]")] \\
2.[("Corfu", "location.location.containedby", "[ENT1]"), ("[ENT1]", \text{\;\;\;\;}"location.country.official\_language", "[ENT2]")] \\
3.[("Corfu", "location.administrative\_division.country", "[ENT1]"), ("[ENT1]", \text{\;\;\;\;}"location.country.official\_language", "[ENT2]")] \\
\end{minipage} \\

\hline

\textbf{Retrieved} & 
\begin{minipage}[t]{\linewidth}
\raggedright
1.[("Corfu", "location.administrative\_division.country", "Greece"), ("Greece", \text{\;\;\;\;}"location.country.languages\_spoken", "Albanian language")] \\
2.[("Corfu", "location.administrative\_division.country", "Greece"), ("Greece", \text{\;\;\;\;}"location.country.official\_language", "Greek Language")] \\
3.[("Corfu", "location.location.containedby", "Corfu Island"), ("Corfu Island", \text{\;\;\;\;}"common.topic.article", "m.0cc3p")]
\end{minipage} \\

\hline
\raggedright
\textbf{Ground Truth (2 items)} & 
Albanian language, Greek Language \\

\hline

\textbf{Output Answer} & 
Albanian and Greek language \\
\hline
\end{tabular}
\caption{Case study C6: Interpretability analysis on the CWQ dataset.}
\label{tab:c6}
\end{table*}

\begin{table*}[htbp]
\centering

\footnotesize
\renewcommand{\arraystretch}{1.4}
\begin{tabular}{p{2.5cm}|p{11.5cm}}
\hline

\textbf{Question} & 
What European Union country is home to the capital city of Brussels? \\
\hline

\textbf{Assertions} & 
\begin{minipage}[t]{\linewidth}
\raggedright
1. ("[ENT1]'s capital city is Brussels", "European Union contains [ENT1].") \\
2. ("The capital cities of [ENT1] are Brussels.", "The European Union is composed of \text{\;\;\;\;}[ENT1].") \\
3. ("Brussels serves as the capital city for [ENT1].", "The member states of the European \text{\;\;\;\;}Union are [ENT1].") \\
4. ("Brussels is the capital city of [ENT1]", "European Union contains [ENT1].")
\end{minipage} \\
\hline

\textbf{Strategy} & 
\begin{minipage}[t]{\linewidth}
Precision
\end{minipage} \\
\hline

\textbf{Schema Graphs} & 
\begin{minipage}[t]{\linewidth}
\raggedright
1. [("Brussels", "location.administrative\_division.capital", "[ENT1]"]), ("[ENT1]", \text{\;\;\;\;}"location.location.containedby", "European Union")] \\
2. [("Brussels", "location.location.containedby", "[ENT1]"]), ("[ENT1]", \text{\;\;\;\;}"location.location.containedby", "European Union")] \\
3. [("Brussels", "location.administrative\_division.capital", "[ENT1]"]), ("[ENT1]", \text{\;\;\;\;}"organization.membership\_organization.members", "European Union")] \\
4. [("Brussels", "location.administrative\_division.capital", "[ENT1]"]), ("[ENT1]", \text{\;\;\;\;}"location.location.containedby", "European Union")] \\
\end{minipage} \\

\hline

\textbf{Retrieved} & 
\begin{minipage}[t]{\linewidth}
\raggedright
1. [("European Union", "organization.organization.founders", "Belgium"), ("Brussels", \text{\;\;\;\;}"location.administrative\_division.capital", "Belgium")] \\
2. [("European Union", "organization.membership\_organization.members", "France"), \text{\;\;\;\;}("Paris", "location.administrative\_division.capital", "France")] \\
\end{minipage} \\

\hline

\textbf{Ground Truth} & 
Belgium \\

\hline

\textbf{Output Answer} & 
Belgium \\
\hline
\end{tabular}
\caption{Case study C7: Interpretability analysis on the CWQ dataset.}
\label{tab:c7}
\end{table*}

\subsection{Efficiency Analysis}
\label{sec:ea}
The STEM framework is designed to minimize reliance on heavy, iterative LLM calls. For a single query, the process involves exactly three distinct LLM inference steps: 
\begin{itemize}
\item\textbf{Projection}: For both the SGDA and SAGB module, we deploy a locally hosted model respectively. So the projection pipeline necessitates a single inference call of the 8B SGDA model to generate \( \mathcal{B} \) planning candidates, followed by \( \mathcal{B} \) forward passes of the 8B SAGB model, the total number of model invocations is \( (2 \times \mathcal{B}) \). Compared to FiDeLiS, which necessitates $(\mathcal{B}×L+L+1)$ LLM calls (where $L$ denotes reasoning depth), our projection operation scales linearly with the beam size while maintaining optimal performance. This efficiency is comparable to existing high-efficiency methods such as LMP $(2L+1)$ \cite{wan-etal-2025-digest} and RoG $(N+1)$, where $N$ represents the number of generated relation paths. Moreover, due to the significantly shorter generation length of our assertions and schema graphs, the actual inference latency is further reduced.

\item\textbf{Triple-Dependent GNN}: We execute a single forward pass of a 6-layer Triple-Dependent GNN to compute the Guidance Graph. Since node and triple embeddings can be pre-indexed, the online computational cost is restricted to the propagation of query-specific interaction scores. 
\item\textbf{Parallel Structure-Tracing Retrieval}: Since the entity selection phase yields multiple entry entities simultaneously, we employ a parallel execution strategy, where search threads originating from all the anchored nodes are conducted concurrently. By parallelizing the retrieval process and merging the resulting subgraphs, we significantly reduce the wall-clock time compared to sequential search methods. Consequently, the overall time cost of subgraph retrieval takes approximately 1.5 to 2 times longer than retrieval from a single entry node.
\item\textbf{Answer Generation}: Once the evidence subgraph is retrieved and linearized, the generator LLM is invoked once to produce the final response. 
\end{itemize}

A critical factor influencing the execution efficiency of STEM is the subgraph search mode, which is determined by the answer strategy. Consequently, we conduct a focused inference analysis of different search mode. All experiments were conducted on a single NVIDIA H100 GPU. 

Given the close correlation between answer number and search modes, we stratify the test samples by answer number and evaluate the inference efficiency for each group independently. We first present the strategy generation accuracy of the SGDA module in Table \ref{tab:sgda_acc}. The results demonstrate that the SGDA module exhibits strong performance in strategy discrimination, achieving an accuracy of over 90\% across all dataset splits.

Subsequently, we present a comparison of inference efficiency grouped by answer counts in Table \ref{tab:per_speed_ans_count}, which indicates that inference latency increases significantly as the number of answers grows. Notably, when the answer count exceeds 10, the inference time rises to over 9 seconds. This is attributed to the fact that a higher volume of answers corresponds to a greater number of matched KG edges aligning with the schema graph.

We further evaluated the test set by stratifying samples based on different answer strategies, with results shown in Table \ref{tab:pre_brd}. Although inference latency is significantly higher in \textit{Breadth} mode than in \textit{Precision} mode, the substantial improvement in F1 score with \textit{Breadth} mode (Section \ref{sec:appendixd1}) justifies this trade-off for more comprehensive answers. We consider the moderate latency increase acceptable, especially given that real-world deployment efficiency is expected to improve with advancing technologies.

\textbf{Pruning Strategy.} The threshold-based search mode incorporates a pruning mechanism to prevent computational explosion. Taking Case C6 (Table \ref{tab:c6}) as an example, the reasoning chain requires traversing from ``Corfu'' to ``[ENT1]'' and subsequently to ``[ENT2]''. The retrieval results reveal that during the transition from ``Corfu'' to ``[ENT1]'', the threshold filtering mechanism retained only a single unique edge, effectively converging to one path. Parallel branching emerged only during the subsequent expansion from ``[ENT1]'' to ``[ENT2]'' to capture all valid answers. This demonstrates that parallel paths are triggered specifically when necessary to cover multiple answers, thereby significantly preserving computational efficiency.

\textbf{Efficiency Comparison with Baselines.} We compare our proposed method against existing baselines to demonstrate its advantages and superiority. To better contextualize the inference latency, we first categorize the existing approaches into three paradigms: (1) Interactive methods involve iterative LLM calls during the retrieval process. (2) Generation-based methods require constant number of LLM calls upfront for path or plan generation. (3) Retrieval-based methods rely entirely on GNNs for graph context encoding. The results are shown in Table \ref{tab:baseline_eff}.

Empirical results demonstrate that STEM is substantially faster than interactive baselines. Specifically, it achieves a nearly a 6-fold speedup over FiDeLiS (5.77s vs. 34.97s) and operates 2.5 times faster than PoG (5.77s vs. 14.28s). This validates our claim that the offline Structure-Tracing paradigm effectively circumvents the computational bottleneck of repetitive LLM reasoning. While STEM is marginally slower than RoG (which generates linear relation paths) and GNN-RAG (which relies on node-level scoring), it delivers vastly superior retrieval accuracy, representing a highly favorable trade-off between efficiency and effectiveness.

\begin{table}[t]
\footnotesize
\centering
\renewcommand{\arraystretch}{1.15}
\setlength{\tabcolsep}{3.5pt} 

\begin{tabular}{l|c|c|c}
\toprule
\hline
{\textbf{Method}} & \textbf{Type} & {\textbf{WebQSP}} & {\textbf{CWQ}} \\

\hline
\textbf{FiDeLiS} & Interactive	 &  34.97	 &  40.16 \\
\textbf{PoG} & Interactive & 14.28 & 14.08 \\
\textbf{RoG} & 	Generation &	3.75	& 4.31 \\
\textbf{GNN-RAG} & 	Retrieval	& 2.64	& 3.81 \\
\textbf{STEM (Ours)}	& Generation	& 5.77	& 5.40 \\
\hline
\bottomrule
\end{tabular}
\caption{Comparison of average inference latency across different methods. Values denote the average wall-clock time (s) required to process a single query.}
\label{tab:baseline_eff}
\end{table}

\renewcommand{\algorithmicrequire}{\textbf{Input:}}
\renewcommand{\algorithmicensure}{\textbf{Output:}}

\subsection{Interpretability Analysis}
STEM offers a transparent, interpretable workflow rooted in two key structural mechanisms:
\begin{itemize}
    \item \textbf{Explicit Logical Blueprinting}: By leveraging the linguistic and reasoning capabilities of LLMs within a KG-constrained framework, STEM explicitly visualizes the reasoning process. The SGDA transforms the ambiguity of natural language into a sequence of logical atomic relational assertions, which the SAGB then projects into a concrete schema graph. The resulting blueprint is not merely a semantic abstraction but a topologically valid plan aligned with the latent knowledge structures of the target KG. This allows us to directly inspect and verify the model's decomposition logic before any retrieval occurs. 
    \item \textbf{Guided Structural Tracing}: Building on this blueprint, the Structure-Tracing Retrieval mechanism enables global matching based on the logical subgraph structure, directly mapping the query's structural features onto the KG knowledge. The schema graph acts as a navigational chart, while the Guidance Graph derived from the Triple-GNN serves as a soft structural bias. This dual-guidance system ensures that the search process is not driven solely by one-sided semantic matching—which is prone to deviation—but is anchored by global structural hypotheses. 
\end{itemize}

\section{Analysis of Failure Modes and Error Propagation}
\label{sec:afmep}

Given the sequential architecture of STEM—where the outputs of the SGDA and SAGB modules serve as inputs for the Triple-GNN and subsequent retrieval—inaccuracies at any upstream stage can inevitably propagate downstream. To better understand this cascading effect, this section presents a systematic analysis of failure modes and error accumulation across the STEM pipeline. 

Specifically, we construct two error attribution matrices on WebQSP to reveal how errors at different stages affect the final QA performance. These matrices capture the cascading effects of two critical intermediate phases: (1) Schema Generation Correctness (SGDA \& SAGB) versus Final QA Accuracy, and (2) Retrieved Evidence Subgraph Correctness versus Final QA Accuracy. We provide a detailed analysis of each dimension below.

\textbf{Schema Generation Correctness vs. QA Accuracy}: We constructed an error attribution matrix to evaluate the correlation between Schema Generation Correctness (produced by the SGDA and SAGB pipeline) and the final QA Accuracy. The results are presented in Table \ref{tab:sgcqm}.

The results show that the highest proportion (85.24\%) occurs when both planning and QA are correct, demonstrating the effectiveness of the query planning modules. The 8.67\% where planning is correct but QA fails likely stems from retrieval errors or the LLM not utilizing the correct evidence. In 5.91\% of cases, planning fails yet QA succeeds—almost entirely attributable to the LLM’s parametric knowledge. The matrix reveals that incorrect schema graphs account for only 16.4\% of total QA failures (0.017/0.1037). This indicates that the vast majority of errors originate in downstream stages—namely, during subgraph retrieval and the LLM's final answer generation—rather than from upstream planning.

\textbf{Retrieved Evidence Subgraph Correctness vs. QA Accuracy}: We analyze the correlation between retrieval correctness and final answer correctness. The results are shown in Table \ref{tab:rescq}.

The proportion of cases where retrieval is correct and the answer is correct reaches 83.01\%, reflecting the effectiveness of the retrieval algorithm and its contribution to question answering. In 4.12\% of cases, retrieval is correct but the answer is incorrect, which is primarily attributable to the LLM's failure in contextual answer extraction. Cases where retrieval is incorrect yet the answer is correct account for 7.34\%, largely due to the LLM's parametric knowledge. Finally, instances where both retrieval and QA fail constitute 5.53\%. From this analysis, we consider the 83.01\%—where both retrieval and QA are correct—as the true reflection of STEM's capability, demonstrating the robustness and accuracy of its retrieval mechanism and its positive impact on final answer generation. Further analysis reveals that retrieval failures account for 57.3\% (0.0553/0.0965) of all incorrect answers. This indicates that flawed evidence retrieval is the primary source of downstream errors, with the remaining 42.7\% attributable to inherent LLM hallucinations during the final generation phase.

\begin{table}[t]
\centering
\renewcommand{\arraystretch}{1.15}
\setlength{\tabcolsep}{6pt}

\begin{tabular}{c|cc}
\hline
\multirow{2}{*}{\( \mathcal{G}_{sch} \) \textbf{match}} & \multicolumn{2}{c}{\textbf{Final QA Output}} \\ \cline{2-3}
& Correct & Incorrect \\
\hline
\textbf{Valid} & 85.24 & 8.67 \\
\textbf{Invalid}    & 5.91 & 1.7 \\
\hline
\end{tabular}
\caption{Schema Generation Correctness vs. QA Accuracy Matrix. Rows indicate whether the schema graph produced by the SGDA and SAGB pipeline successfully matches the ground-truth reasoning graph, while columns represent whether the final generated answer is correct. All values are reported as percentages (\%).}
\label{tab:sgcqm}
\end{table}

\begin{table}[t]
\centering
\renewcommand{\arraystretch}{1.15}
\setlength{\tabcolsep}{6pt}

\begin{tabular}{c|cc}
\hline
\multirow{2}{*}{$\mathcal{G}_{\text{reason}}$ \textbf{match}} & \multicolumn{2}{c}{\textbf{Final QA Output}} \\ \cline{2-3}
& Correct & Incorrect \\
\hline
\textbf{Valid} & 83.01 & 4.12 \\
\textbf{Invalid}    & 7.34 & 5.53 \\
\hline
\end{tabular}
\caption{Retrieved Evidence Subgraph Correctness vs. QA Accuracy. The rows correspond to the correctness of the retrieved evidence subgraph, while the columns represent the correctness of the final answer. All values are reported as percentages (\%).}
\label{tab:rescq}
\end{table}

\section{Detailed Mechanism of the Semantic-to-Structural Projection and Structural Pattern Acquisition}
\label{sec:projection_pipeline}

To address potential ambiguities regarding how our method maps a natural language text to schema graph, we provide a detailed walkthrough of the Semantic-to-Structural Projection pipeline. This pipeline, comprising the SGDA and SAGB modules, operates as an end-to-end generative translation process, converting complex raw queries into a set of triples that constitute the schema graph.

To illustrate this workflow concretely, consider the query: ``where is the fukushima daiichi nuclear plant located'', it's processed through the following two stages:

\paragraph{Stage 1: SGDA Decomposition} 
The SGDA module first decomposes the raw text query into a set of atomic relational assertions, ultimately yielding:
\begin{align*}
&\text{"The fukushima daiichi nuclear power plant } \\
&\text{is contained by [ENT1]."}
\end{align*}
Here, [ENT1] serves as a structural placeholder for the unknown answer or intermediate entity, simplifying the subsequent mapping task.

\paragraph{Stage 2: SAGB Alignment} 
To perform the symbolic mapping, the SAGB module autoregressively translates the input atomic assertions into the corresponding valid KG triples:
\begin{align*}
&\text{("The fukushima daiichi nuclear power plant", } \\
&\text{"location.location.containedby",} \\
&\text{"[ENT1]")}
\end{align*}

This process successfully translates highly variable natural language into strict KG-oriented structures by leveraging the LLM's reasoning capabilities. Since this structural generalization capability is explicitly instilled through our training regimen, we next provide concrete examples to illustrate how it is achieved.

\paragraph{Generalization over Training}
Importantly, this Semantic-to-Structural Projection learns underlying alignment patterns rather than merely memorizing specific facts. During the fine-tuning phase, the training data for SGDA and SAGB contain structurally similar alignments, such as those illustrated in Figures \ref{fig:t1} and \ref{fig:t2}.

\begin{figure}[H]
    \centering
    \footnotesize
    \begin{tcolorbox}[
        title={Example (1)},      % 标题
        colback=gray!5,                        % 背景色
        colframe=gray!60,                  % 边框色
        coltitle=white,                       % 标题文字色
        fonttitle=\small\bfseries,            % 标题字体
        boxrule=0.8pt,                        % 边框粗细
        left=2mm, right=2mm, top=2mm, bottom=2mm, % 内边距
        arc=1mm,                              % 圆角
        width=\columnwidth                      % 【关键】宽度设为全页宽度
    ]
    
    \textbf{Query:} In which country is Kagoshima Prefecture located? \\
    \textbf{Atomic Relational Assertions:} ("Kagoshima Prefecture is contained by [ENT1]",) \\
    \textbf{Schema Graph:} [("Kagoshima Prefecture",  
    
    "location.location.containedby", "[ENT1]")] \\
    \textbf{Answer:} Japan
    \end{tcolorbox}
    \caption{SDGA \& SAGB Training Data Example (1).}
    \label{fig:t1}
\end{figure}

\begin{figure}[H]
    \centering
    \footnotesize
    \begin{tcolorbox}[
        title={Example (2)},      % 标题
        colback=gray!5,                        % 背景色
        colframe=gray!60,                  % 边框色
        coltitle=white,                       % 标题文字色
        fonttitle=\small\bfseries,            % 标题字体
        boxrule=0.8pt,                        % 边框粗细
        left=2mm, right=2mm, top=2mm, bottom=2mm, % 内边距
        arc=1mm,                              % 圆角
        width=\columnwidth                      % 【关键】宽度设为全页宽度
    ]
    
    \textbf{Query:} Which city in Aomori Prefecture was affected by the 2011 Tohoku earthquake? \\
    \textbf{Atomic Relational Assertions:} ("[ENT1] is contained by Aomori Prefecture.", "[ENT1] experienced the event of the 2011 Tōhoku earthquake and tsunami") \\
    \textbf{Schema Graph:} [("[ENT1]",  \\
    "location.location.containedby", "Aomori Prefecture"), \\
    ("[ENT1]", "location.location.events", \\
    "2011 Tōhoku earthquake and tsunami")]  \\
    \textbf{Answer:} Tohoku
    \end{tcolorbox}
    \caption{SDGA \& SAGB Training Data Example (2).}
    \label{fig:t2}
\end{figure}

After capturing these schema patterns, the pipeline can effectively generalize to structurally similar assertions (e.g., ``X is located in Y'') across different entities. This generative design enables STEM to perform robust, structure-aware schema alignment, circumventing the rigidity and out-of-vocabulary issues typical of traditional step-wise path search or dictionary-based matching methods.

\section{Prompt List}
\label{sec:promptlist}

To ensure the robustness and reproducibility of STEM, we detail the core prompts utilized across the different stages of our pipeline. These encompass the processing prompts for the SGDA and SAGB modules, the generation prompt for the QA model, the assertion synthesis prompt for training data construction, the prompt for determining the response strategy, and notably, the core feature of our work: the Structure-to-Query Reverse Generation data synthesis prompt, which significantly enhances the model's structural generalization capabilities. We present the complete instructional content of each prompt in Figures \ref{fig:p1} through \ref{fig:p6}.

\definecolor{headergray}{RGB}{160,160,160}

\subsection{Schema-Aligned Question Decomposition Prompt \hyperref[fig:p1]{$ \mathcal{P}_1 $}}
\begin{figure}[H]
    \centering
    \footnotesize
    \begin{tcolorbox}[
        title={Schema-Aligned Question Decomposition Prompt ($\mathcal{P}_1$)},      % 标题
        colback=gray!5,                        % 背景色
        colframe=gray!60,                  % 边框色
        coltitle=white,                       % 标题文字色
        fonttitle=\small\bfseries,            % 标题字体
        boxrule=0.8pt,                        % 边框粗细
        left=2mm, right=2mm, top=2mm, bottom=2mm, % 内边距
        arc=1mm,                              % 圆角
        width=\columnwidth                      % 【关键】宽度设为全页宽度
    ]
    
    You are a multi-hop question decomposition expert specialized in knowledge graph-based question answering. \\ Your task is to decompose an input multi-hop question into a sequence of single-hop assertions, and returns the answer strategy required to answer the query. The specific requirements are as follows:
    \begin{itemize}
    \item 1. Decomposition must be entity-centric. Each single-hop assertion should correspond to a pair of entities and describe the relationship between them. 
    \item 2. Every single-hop assertion generated should contribute to answering the original multi-hop question.
    \item 3. For entity references—including the final answer or any intermediate answer entities—you must label them with [ENT1], [ENT2], etc., and ensure that: \\
    \text{      }· The same entity is consistently referred to with the same label across all single-hop assertions.\\
    \text{      }· Different entities are assigned distinct labels. 
    \item 4. Answering Strategy: If you determine that the answer to the current question is exclusive and deterministic, please return the strategy ``Precision''. If you assess that the question involves multiple distinct answers (including final or intermediate answers), please return the strategy ``Breadth''. You must select your strategy strictly from ``Precision'' and ``Breadth''. \\
    \end{itemize}
    Your response should be a single result consisting of the planned single-hop assertion (s) along with the strategy. \\
    \textbf{Your output format}: \\
    \text{(Assertion\_1, Assertion\_2, Assertion\_3), Strategy)} \

    \vspace{0.3cm}
    
    \textbf{Example:} \\
    \text{[EXAMPLE\_1]} \\
    \text{[EXAMPLE\_2]} \\
    \text{[EXAMPLE\_3]} \\
    
    \vspace{0.3cm}
    
    \textbf{Input multi‑hop question:} \\
    \text{[Query]} \\
    Please now plan the decomposition for the given question. You must strictly follow the requirements above.
    
    \end{tcolorbox}
    % 如果需要图注，取消下面这行的注释
    \caption{The prompt template for Schema-Aligned Question Decomposition ($\mathcal{P}_1$).}
    \label{fig:p1}
\end{figure}

\text{} \\
\text{} \\
\text{} \\
\text{} \\
\text{} \\
\text{}

\subsection{Schema Graph Construction Prompt \hyperref[fig:p2]{$ \mathcal{P}_2 $}}

\begin{figure}[H]
    \centering
    \footnotesize
    \begin{tcolorbox}[
        title={Schema Graph Construction Prompt ($\mathcal{P}_2$)},      % 标题
        colback=gray!5,                        % 背景色
        colframe=gray!60,                  % 边框色
        coltitle=white,                       % 标题文字色
        fonttitle=\small\bfseries,            % 标题字体
        boxrule=0.8pt,                        % 边框粗细
        left=2mm, right=2mm, top=2mm, bottom=2mm, % 内边距
        arc=1mm,                              % 圆角
        width=\columnwidth                      % 【关键】宽度设为全页宽度
    ]
    
    You are an entity-relationship construction expert who has memorized a rich and professional knowledge graph-oriented semantic and logical structure. Based on your mastered graph structure data, you can construct appropriate entity-relationship triples for given single-hop assertions. 
    
    Since these assertions are decomposed from a multi-hop query, you must fully consider their interdependencies when returning the triples. \\
    The specific requirements are as follows:
    \begin{itemize}
    \item 1. Based on the given assertions and combined with your understanding of knowledge graph data, you must generate structural triples that best match the meaning expressed. The entities and relationship descriptions in the triples must be consistent with the meaning of the assertions.
    \item 2. If an assertion contains an entity placeholder like [ENTX], you must copy it exactly as is when converting it into a triple. Do not alter the content of the placeholder. If there is no [ENTX] label in an assertion, generate the most relevant triple based on the described entities and relationship.
    \end{itemize}
  
    \textbf{Your output format}: \\
    \text{[}\\
    \text{      } (Entity\_1, Relation\_1, Entity\_2), \\
    \text{      } (Entity\_2, Relation\_2, Entity\_3), \\
    \text{      } ...\\
    \text{]}

    \vspace{0.3cm}
    
    \textbf{Example:} \\
    \text{[EXAMPLE\_1]} \\
    \text{[EXAMPLE\_2]} \\
    
    \vspace{0.3cm}
    
    \textbf{Input single-hop assertions:} \\
    \text{[Assertion\_1, Assertion\_2, ...]}
    
    \end{tcolorbox}
    % 如果需要图注，取消下面这行的注释
    \caption{The prompt template for Schema Graph Construction ($\mathcal{P}_2$).}
    \label{fig:p2}
\end{figure}

\subsection{Generation Prompt \hyperref[fig:p3]{$ \mathcal{P}_3 $}}

\begin{figure}[H]
    \centering
    \footnotesize
    \begin{tcolorbox}[
        title={Generation Prompt ($\mathcal{P}_3$)},      % 标题
        colback=gray!5,                        % 背景色
        colframe=gray!60,                  % 边框色
        coltitle=white,                       % 标题文字色
        fonttitle=\small\bfseries,            % 标题字体
        boxrule=0.8pt,                        % 边框粗细
        left=2mm, right=2mm, top=2mm, bottom=2mm, % 内边距
        arc=1mm,                              % 圆角
        width=\columnwidth                      % 【关键】宽度设为全页宽度
    ]
    
    Based on the knowledge structure graph, please answer the
    given question. Please keep the answer as simple as possible
    and return all the possible answers as a list.

    \vspace{0.3cm}

    \textbf{Knowledge Structure Graph}: [Knowledge Structure Graph]\\
    \textbf{Question}: [Question]\\
    \textbf{Answer}: [Answer]\\
    \end{tcolorbox}
    \caption{The prompt template for Generation ($\mathcal{P}_3$).}
    \label{fig:p3}
\end{figure}

\subsection{Path-based Assertions Generation Prompt \hyperref[fig:p4]{$ \mathcal{P}_4 $}}

\begin{figure}[H]
    \centering
    \footnotesize
    \begin{tcolorbox}[
        title={Path-based Assertions Generation Prompt ($\mathcal{P}_4$)},      % 标题
        colback=gray!5,                        % 背景色
        colframe=gray!60,                  % 边框色
        coltitle=white,                       % 标题文字色
        fonttitle=\small\bfseries,            % 标题字体
        boxrule=0.8pt,                        % 边框粗细
        left=2mm, right=2mm, top=2mm, bottom=2mm, % 内边距
        arc=1mm,                              % 圆角
        width=\columnwidth                      % 【关键】宽度设为全页宽度
    ]
    
    Please construct appropriate declarative assertions for each given triple based on the provided logical triple list and original query. The requirements are as follows: 
    \begin{itemize}
    \item 1. The format of each given triple is (``entity1'', ``relation'', ``entity2''). You need to generate a suitable declarative assertion based on the meanings of entity1 and entity2 and the relationship between them.
    \item 2. If an entity is in the label format such as [ENTX], the corresponding entity in the generated sentence should also be written in the same label format. If multiple triples contain the same label, for example, all are [ENT1], you must ensure consistency in the generated assertions and avoid modifying the label content arbitrarily.
    \item 3. The number of assertions you return must match the number of triples in the given list, and they should correspond one-to-one.
    \item 4. The assertions you generate must serve as important evidence for answering the given original query, meaning that answering the query requires referencing these assertions.
    \end{itemize}

    \textbf{Your output format}: \\
    \text{[}\\
    \text{      } Assertion\_1, \\
    \text{      } Assertion\_2, \\
    \text{      } ...\\
    \text{]}

    \textbf{Example:} \\
    \text{[EXAMPLE\_1]} \\
    \text{[EXAMPLE\_2]} \\
    \text{[EXAMPLE\_3]} \\

    \vspace{0.3cm}

    \textbf{Original Query}: \\
    \text{[Query]}

    \textbf{Given Triple List}: \\
    \text{[Triple\_1, Triple\_2, ...]}
    \end{tcolorbox}
    \caption{The prompt template for Path-based Assertions Generation ($\mathcal{P}_4$).}
    \label{fig:p4}
\end{figure}

\text{} \\
\text{} \\
\text{} \\
\text{} \\
\text{} \\
\text{} \\
\text{} \\
\text{} \\
\text{} \\
\text{}

\subsection{Response Strategy Generation Prompt \hyperref[fig:p5]{$ \mathcal{P}_5 $}}

\begin{figure}[H]
    \centering
    \footnotesize
    \begin{tcolorbox}[
        title={Response Strategy Generation Prompt ($\mathcal{P}_5$)},      % 标题
        colback=gray!5,                        % 背景色
        colframe=gray!60,                  % 边框色
        coltitle=white,                       % 标题文字色
        fonttitle=\small\bfseries,            % 标题字体
        boxrule=0.8pt,                        % 边框粗细
        left=2mm, right=2mm, top=2mm, bottom=2mm, % 内边距
        arc=1mm,                              % 圆角
        width=\columnwidth                      % 【关键】宽度设为全页宽度
    ]
    
    Please determine the appropriate retrieval strategy for the given question when used for multi-hop retrieval-augmented generation in a knowledge graph context. There are two available strategies: ``Precision'' and ``Breadth'', described as follows: \\
    \begin{itemize}
    
    \item 1. ``Precision'' is primarily used for answering questions that have a exclusive and
    deterministic answer, such as ``In which year was Edison born?'' Questions of this type typically have only one correct answer, and contradictions may arise if more than two answers are provided.
    \item 2. ``Breadth'' is mainly used for answering questions involving multiple distinct answers, such as ``What country is Russia close to?'', this type of question requires retrieving all qualifying answers to ensure comprehensive answer recall.
    \item 3. You will be given a list of assertions representing the decomposed declarative statements derived from the given question. These assertions constitute the multi-hop logical decomposition of the question. You may reference them to get more in-depth guidance.
    \end{itemize}

    Please remember, your response should contain only the word ``Precision'' or ``Breadth'', with no additional explanatory content!
    \vspace{0.3cm}

    \textbf{Example:} \\
    \text{[EXAMPLE\_1]} \\
    \text{[EXAMPLE\_2]} \\
    \text{[EXAMPLE\_3]} \\

    \vspace{0.3cm}

    \textbf{Question}: [Question] \\
    \textbf{Assertions}: [Assertion\_1, Assertion\_2, ...]
    \end{tcolorbox}
    \caption{The prompt template for Response Strategy Generation ($\mathcal{P}_5$).}
    \label{fig:p5}
\end{figure}

\text{} \\
\text{} \\
\text{} \\
\text{} \\
\text{} \\
\text{} \\
\text{} \\
\text{} \\
\text{} \\
\text{} \\
\text{} \\
\text{}

\subsection{Query \& Assertions Generation based on Sampled-Graph Prompt \hyperref[fig:p6]{$ \mathcal{P}_6 $}}

\begin{figure}[H]
    \centering
    \footnotesize
    \begin{tcolorbox}[
        title={Query \& Assertions Generation based on Sampled-Graph Prompt ($\mathcal{P}_6)$},      colback=gray!5,                        % 背景色
        colframe=gray!60,                  % 边框色
        coltitle=white,                       % 标题文字色
        fonttitle=\small\bfseries,            % 标题字体
        boxrule=0.8pt,                        % 边框粗细
        left=2mm, right=2mm, top=2mm, bottom=2mm, % 内边距
        arc=1mm,                              % 圆角
        width=\columnwidth                      % 【关键】宽度设为全页宽度
    ]
    
    Please construct appropriate declarative sentences for each given triple based on the provided triple list. The requirements are as follows: 
    \begin{itemize}
    \item 1. The format of each given triple is (``entity1'', ``relation'', ``entity2''). You need to generate a suitable declarative sentence based on the meanings of entity1 and entity2 and the relationship between them.
    \item 2. If an entity is in the label format such as [ENTX], the corresponding entity in the generated sentence should also be written in the same label format. If multiple triples contain the same label, you must ensure consistency in the generated sentences and avoid modifying the label content arbitrarily.
    \item 3. The number of sentences you return must match the number of triples in the given list, and they should correspond one-to-one.
    \end{itemize}

    After constructing the declarative sentences, proceed to generate a multi-hop question based on them. You must adhere to the following requirements:
    \begin{itemize}
    \item 1. Given the specific placeholder [ENTX] designated as the answer entity, your generated question must target [ENTX] as its final answer.
    \item 2. If the sentences contain multiple distinct placeholders, treat all placeholders other than the target [ENTX] as intermediate answers. You must integrate them into the multi-hop question as modifiers, relative clauses, or nested constraints.
    \item 3. The generated multi-hop question must be constructed such that answering it requires referencing all the provided declarative sentences.
    \item 4. If you determine that the declarative sentences cannot yield an unambiguous multi-hop question that strictly relies on every sentence, simply output: ``No Solution''.
    \end{itemize}

    \textbf{Your output format}: \\
    \texttt{[(Sentence\_1, Sentence\_2, ...), Multi-Hop Question]}

    \vspace{0.3cm}

    \textbf{Example:} \\
    \text{[EXAMPLE\_1]} \\
    \text{[EXAMPLE\_2]} \\
    \text{[EXAMPLE\_3]} \\

    \vspace{0.3cm}

    \textbf{Given Triple List}: [Triple\_1, Triple\_2, ...]\\
    \textbf{Answer Entity}: [ENTX]
    \end{tcolorbox}
    \caption{The prompt template for Query \& Assertions Generation based on Sampled-Graph ($\mathcal{P}_6$).}
    \label{fig:p6}
\end{figure}

\section{Structure-Tracing Subgraph Retrieval}
\label{sec:sstr}
Algorithm \ref{alg:stsr} and \ref{alg:search_func} illustrate the overall execution and helper functions of Structure-Tracing Subgraph Retrieval, encompassing the execution procedures for two search modes. Descriptions of the key helper functions are provided below:
\begin{itemize}
\item \textsc{Contradict}: Determines whether a specific triple $t$ already exists within the list of currently matched triples.
\item \textsc{Get\_Tail}: Given a triple $t$ and one of its constituent entities $e$, retrieves the complementary entity node.
\item \textsc{GET\_n}: Retrieves all incident edges (not differentiate edge direction) of entity $e$ within and $\mathcal{G}$ returns them as a list of triples.
\item \textsc{BUILD\_GRAPH}: Constructs a graph structure from a list of triples.
\end{itemize}

% 注意：在双栏模板中，跨栏环境通常只能放在页面的顶部 [t] 或底部 [b]
\begin{algorithm*}[t] 
\caption{Structure-Tracing Subgraph Retrieval}
\label{alg:stsr}
\small
\begin{algorithmic}[1]

% --- Input / Output (手动分行，防止一行太长) ---
\State \textbf{Input:} Pattern graph $\mathcal{G}_{sch} = (\mathcal{N}_Q, \mathcal{R}_Q)$; Knowledge Graph $\mathcal{G}$; Topic entity $e \in \mathcal{G}_{sch}$; 
\Statex \hskip 2.6em Globally-aware entity score $S^*_e$; Retrieval strategy $\sigma$; Confidence threshold $\theta$.
\State \textbf{Output:} Matched subgraph $\mathcal{G}^*_{sch}$.

\Statex \hrulefill \Statex % 一条细横线，美观

% --- 主逻辑区 ---
\State $e^* \gets \text{Identify the matching entity node in } \mathcal{G} \text{ corresponding to } e \text{ in } \mathcal{G}_{sch}$
\State $S_0 \gets S^*_e[e^*]; \quad \mathit{final\_list} \gets \emptyset; \quad \mathit{last\_visit} \gets \text{None}$ \Comment{Initialization}

\State $\mathcal{T}_{all} \gets \textsc{Match}(\mathcal{G}, \mathcal{G}_{sch}, S_0, e, e^*, \mathit{final\_list}, \mathit{last\_visit}, \sigma, \theta)$
\State $\mathcal{G}^*_{sch} \gets \textsc{Build\_Graph}(\mathcal{T}_{all})$
\State \textbf{return} $\mathcal{G}^*_{sch}$

\Statex 

% --- 辅助函数定义 ---
\Function{Contradict}{$t, \mathit{final\_list}$}
    \If{$t \in \mathit{final\_list}$} \Return \textsc{True} \Else \ \Return \textsc{False} \EndIf
\EndFunction

\Statex 

\Function{Get\_n}{$e, \mathcal{G}$}
    \State $\mathit{triples} \gets \text{Fetch all incident edges for entity } e \text{ in } \mathcal{G} \text{ as a triple list}$
    \State \Return $\mathit{triples}$
\EndFunction

\Statex 

\Function{Match}{$\mathcal{G}, \mathcal{G}_{sch}, S, e, e^*, \mathit{final\_list}, \mathit{last\_visit}, \sigma, \theta$}
    \For{\textbf{each} $t \in \textsc{Get\_n}(e, \mathcal{G}_{sch})$ }
        \If{$\mathit{last\_visit} = t$} \textbf{continue} \EndIf
        
        \If{$\sigma = \text{"Precision"}$}
            \State $\mathit{step\_list} \gets \textsc{Step\_Precision}(\mathcal{G}, \mathcal{G}_{sch}, S, e, e^*, t, \mathit{final\_list}, \sigma)$
        \EndIf
        
        \If{$\sigma = \text{"Breadth"}$}
            \State $\mathit{step\_list} \gets \textsc{Step\_Breadth}(\mathcal{G}, \mathcal{G}_{sch}, S, e, e^*, t, \mathit{final\_list}, \sigma, \theta)$
        \EndIf
        \State $\mathit{final\_list} \gets \mathit{final\_list} \cup \mathit{step\_list}$
    \EndFor
    \State \Return $\mathit{final\_list}$
\EndFunction

\end{algorithmic}
\end{algorithm*}

\begin{algorithm*}[t]
\caption{Subgraph Search Single Step Functions}
\label{alg:search_func}
\small
\begin{algorithmic}[1]

% --- STEP_PRECISION ---
\State \textbf{function} \textsc{Step\_Precision}($\mathcal{G}, \mathcal{G}_{sch}, S, e, e^*, t, \mathit{final\_list}, \sigma$)
\State \quad $N_{e^*} \gets \textsc{Get\_n}(e^*, \mathcal{G})$
\State \quad $\mathit{max\_score} \gets -1$
\State \quad $\mathit{max\_t} \gets \text{None}$
\State \quad \textbf{for each} $t^*$ \text{in} $N_{e^*}$ \textbf{do}
\State \quad \quad \textbf{if} \textsc{Contradict}($t^*, \mathit{final\_list}$) \textbf{then}
\State \quad \quad \quad \textbf{continue}
\State \quad \quad $S' \gets S + \textsc{T-Score}(t^*, t)$
\State \quad \quad \textbf{if} $S' > \mathit{max\_score}$ \textbf{then}
\State \quad \quad \quad $\mathit{max\_score} \gets S'$
\State \quad \quad \quad $\mathit{max\_t} \gets t^*$
\State \quad $\mathit{final\_list} \gets \mathit{final\_list} \cup \{\mathit{max\_t}\}$
\State \quad $e_t \gets \textsc{Get\_Tail}(t, e)$
\State \quad \textbf{if} $\mathit{max\_t} \neq \text{None}$ \textbf{then}
\State \quad \quad $e^*_t \gets \textsc{Get\_Tail}(\mathit{max\_t}, e^*)$
\State \quad \quad $\mathit{step\_list} \gets \textsc{Match}(\mathcal{G}, \mathcal{G}_{sch}, \mathit{max\_score}, e_t, \mathit{final\_list}, t, \sigma, \text{None})$
\State \quad \quad $\mathit{final\_list} \gets \mathit{final\_list} \cup \mathit{step\_list}$
\State \quad \textbf{return} $\mathit{final\_list}$

\Statex % 函数间的空行

% --- STEP_BREADTH ---
\State \textbf{function} \textsc{Step\_Breadth}($\mathcal{G}, \mathcal{G}_{sch}, S, e, e^*, t, \mathit{final\_list}, \sigma, \theta$)
\State \quad $N_{e^*} \gets \textsc{Get\_n}(e^*, \mathcal{G})$
\State \quad $\mathit{candidates} \gets \emptyset$
\State \quad \textbf{for each} $t^*$ \text{in} $N_{e^*}$ \textbf{do}
\State \quad \quad \textbf{if} \textsc{Contradict}($t^*, \mathit{final\_list}$) \textbf{then}
\State \quad \quad \quad \textbf{continue}
\State \quad \quad \textbf{if} \textsc{T-Score}($t^*, t) \geq \theta$ \textbf{then}
\State \quad \quad \quad $S' \gets S + \textsc{T-Score}(t^*, t)$
\State \quad \quad \quad $\mathit{candidates} \gets \mathit{candidates} \cup \{(S', t^*)\}$
\State \quad \quad \quad $\mathit{final\_list} \gets \mathit{final\_list} \cup \{t^*\}$
\State \quad $e_t \gets \textsc{Get\_Tail}(t, e)$
\State \quad \textbf{for each} $(\mathit{score}, t^*)$ \text{in} $\mathit{candidates}$ \textbf{do}
\State \quad \quad $e^*_t \gets \textsc{Get\_Tail}(t^*, e^*)$
\State \quad \quad $\mathit{step\_list} \gets \textsc{Match}(\mathcal{G}, \mathcal{G}_{sch}, \mathit{score}, e_t, e^*_t,\mathit{final\_list}, t, \sigma, \theta)$
\State \quad \quad $\mathit{final\_list} \gets \mathit{final\_list} \cup \mathit{step\_list}$
\State \quad \textbf{return} $\mathit{final\_list}$

\end{algorithmic}
\end{algorithm*}

\end{document}